\documentclass[letterpaper]{article} 
\usepackage{JMtwocolumn}  
\usepackage{times}  
\usepackage{helvet} 
\usepackage{courier}  
\usepackage[hyphens]{url}  
\usepackage{graphicx} 
\usepackage{natbib}  
\usepackage{caption} 
\usepackage{amsmath,mathrsfs,amssymb} 
\usepackage{amsfonts}  
\usepackage{algorithm}
\usepackage{algorithmic}
\usepackage{siunitx}   
\usepackage{upgreek}   
\usepackage{xcolor}    
\usepackage{subfigure}
\usepackage{booktabs}       
\usepackage{longtable}
\usepackage{threeparttable}
\usepackage{multirow}
\usepackage{multicol}
\usepackage{times}

\newtheorem{theorem}{Theorem}
\newtheorem{lemma}{Lemma}
\newtheorem{definition}{Definition}

\newtheorem{proposition}{Proposition}

\newcommand*{\dif}{\mathop{}\!\mathrm{d}}

\title{Towards Understanding Theoretical Advantages of Complex-Reaction Networks}
\author{
    Shao-Qun Zhang\textsuperscript{\rm 1}, 
    Wei Gao\textsuperscript{\rm 1}, 
    Zhi-Hua Zhou\textsuperscript{\rm 1} \\
}
\affiliations{
    \textsuperscript{\rm 1} National Key Laboratory for Novel Software Technology \\
    Nanjing University, Nanjing 210023, China \\
    \textit{\{zhangsq,gaow,zhouzh\}@lamda.nju.edu.cn}
}

\begin{document}

\maketitle

\begin{abstract}
	Complex-valued neural networks have attracted increasing attention in recent years, while it remains open on the advantages of complex-valued neural networks in comparison with real-valued networks. This work takes one step on this direction by introducing the \emph{complex-reaction network} with fully-connected feed-forward architecture. We prove the universal approximation property for complex-reaction networks, and show that a class of radial functions can be approximated by a complex-reaction network using the polynomial number of parameters, whereas real-valued networks need at least exponential parameters to reach the same approximation level. For empirical risk minimization, our theoretical result shows that the critical point set of complex-reaction networks is a proper subset of that of real-valued networks, which may show some insights on finding the optimal solutions more easily for complex-reaction networks.
\end{abstract}

\section{Introduction} \label{sec:intro}
Deep neural networks have become a mainstream model of deep learning~\cite{lecun2015} during the past decades, mostly working with real-valued neural networks. For example, great progresses have been made for real-valued neural networks in many real applications such as computer vision \cite{krizhevsky2012}, speech recognition \cite{graves2013,sutskever2014}, machine translation~\cite{bahdanau2014}, etc. Theoretical studies have also attracted much attention on the deep understanding of real-valued neural networks, including universal approximation \cite{barron1994,funahashi1989,hornik1991,kidger2020,leshno1993,lu2017,sun2016}, optimization dynamics \cite{allen2019,dauphin2014,du2019,poggio2020}, out-of-sample generalization~\cite{hardt2016,zhang2020}, neural tangent kernel \cite{jacot2018,du2018,arora2019}, etc.

Recent years have also witnessed an increasing interest on complex-valued neural networks.  Hirose and Yoshida \cite{hirose2012} introduced the complex-valued neural networks with amplitude-phase-type activation function, and showed better generalization than real-valued networks in the fitting interpolation of temporal signals. Mark et al. \cite{tygert2016} simulated the complex-valued convolution operations from the perspective of wavelets. Danihelka et al. \cite{danihelka2016} presented faster learning by adding some complex-valued modules to the recurrent architecture. Some studies showed that complex-valued neural networks can surpass their real-valued contenders in some applications, such as vision \cite{oyallon2015,virtue2017,worrall2017}, NLP \cite{trouillon2016}, signal processing \cite{adali2011,hirose2011,hirose2012}, MRI fingerprinting \cite{virtue2017}, time series forecasting~\cite{wolter2018,zhang2020}, etc.

From the theoretical perspective, Voigtlaender \cite{voigtlaender2020} took an important step on the universal approximation of shallow and deep complex-valued networks, and similar to real-valued networks, complex-valued neural networks could achieve universal approximation with exponential depth or width \cite{arena1993,arena1995,voigtlaender2020}. Several researchers made efforts on optimization dynamics, e.g., all critical points are proven to be saddle points, which are generated from the hierarchical structure of complex-valued networks~\cite{nitta2002,nitta2013}, and Adali et al. \cite{adali2011} showed that complex-valued networks have no bad local minima as for fitting low-degree polynomials, whereas this does not hold for real-valued networks.

This work presents theoretical understandings on the advantages of complex-valued networks in comparison with real-valued networks, and the main contributions can be summarized as follows:
\begin{itemize}
	\item We propose the \emph{Complex-Reaction Network} in the fully-connected feed-forward architecture, and show that the complex-reaction network has universal approximation property.
	\item For approximation, we prove that a kind of radial functions can be approximated by a complex-reaction network with a polynomial number of parameters, whereas the real-valued network cannot arrive at the same approximation level even with exponential ($O(C_1(2d+1)e^{C_1(2d)})$) parameters for some constant $C>0$, where $2d$ denotes the input dimension.
	\item For optimization dynamics, we consider the empirical risk minimization based on the standard gradient descent algorithm. Our theoretical result shows that the critical point set of complex-reaction networks is a proper subset of that of real-valued networks, which may shed some insights on finding optimal solutions more easily for complex-reaction networks.
\end{itemize}

The rest of this paper is organized as follows. Section~\ref{sec:pre} introduces the complex-reaction networks. Section~\ref{sec:approximation} presents the approximation analysis of complex-reaction networks. Section~\ref{sec:OD} studies the optimization dynamics of complex-reaction networks. Section~\ref{sec:conclusion} concludes with future work.

\nocite{hirose2003} 
\nocite{scott2012}  
\nocite{abraham1978} 

\section{Complex-Reaction Networks} \label{sec:pre}
In this section, we first introduce \emph{Complex-Reaction Networks} in the fully-connected feed-forward architecture, and then present the universal approximation analysis.

A complex number $z = z_1 + z_2 \boldsymbol{i}$ comprises a real component $z_1$ and an imaginary component $z_2$. The basic building block of a complex-reaction network can be formalized as
\[
\tau: \mathbb{C}^d \to \mathbb{C},\quad \boldsymbol{z} \mapsto \sigma_{cr}(\boldsymbol{w}^T \boldsymbol{z})
\]
where $\boldsymbol{w} \in \mathbb{C}^d$ denote the connection weights and $\sigma_{cr}$ is a complex-valued activation function. In this work, we employ the following zReLU function~\citep{trabelsi2018,zhang2020} as activation $\sigma_{cr}$
\[
\text{zReLU}(z) = \left\{ \begin{aligned}
	z,~~ & \text{if}~~ \theta_z \in [0,\pi/2] \cup [\pi,3\pi/2] ~, \\
	0,~~ & \text{otherwise} ~.
\end{aligned}\right.
\]
Thus, for $\alpha \in \mathbb{R}^+$, we have
\[
\sigma_{cr}(z) = \frac{\partial \sigma_{cr}(z)}{\partial z} z \quad\text{and}\quad \sigma_{cr}(\alpha z) = \alpha \sigma_{cr}(z) .
\]
Generally, we consider the real-valued data including instances and labels, as the works of~\citep{hirose2012,trabelsi2018,wolter2018}. For complex-valued formation, we enable the first $d$ feature maps to represent the real components and the remaining $d$ to record the imaginary ones, which has been implemented by~\citep{trabelsi2018}. We also employ a pure linear connection as the final layer and extract the real part as the outputs. Therefore, we have established the \emph{Complex-Reaction Network}, $f_{CR}:\mathbb{C}^{d} \to \mathbb{R}$.

We present the universal approximation for complex-reaction networks as follows.
\begin{theorem} \label{thm:approximation}
	For zReLU activation function, the complex-reaction network with one-hidden layer $f_{CR}:\mathbb{C}^{d} \to \mathbb{R}$ is a universal approximator for any continuous function $f:\mathbb{R}^{2d} \to \mathbb{R}$, where $\mathbb{C} \cong \mathbb{R}^2$.
\end{theorem}
Theorem~\ref{thm:approximation} shows that complex-reaction networks have the universal approximation property, similarly to that of the real-valued networks. This theorem can be easily derived from \cite[Theorem 1.3]{voigtlaender2020}, where a shallow complex-valued neural network has the universal approximation property, and we omit the detailed proof of Theorem~\ref{thm:approximation}.

A fully-connected complex-reaction network with one-hidden layer has $l(d+m)$ complex-valued connection weights, which is equivalent to $2l(d+m)$ real-valued connection weights, where $l$ and $m$ denote the number of neurons in hidden and output layers, respectively. Notice that we focus on the number of ``real-valued'' parameters when one mentions the number of parameters in complex-reaction networks. Let $z=z_1 + z_2 \boldsymbol{i}$ be a complex number for $z_1, z_2 \in \mathbb{R}$, and we denote by $\bar{z} = z_1 - z_ 2 \boldsymbol{i}$ and $|z|^2 = z_1^2 + z_2^2$. Let $[\cdot]_R$ and $[\cdot]_I$ denote the operators on the extraction of real and imaginary parts from a complex-valued formation, respectively, for examples, $[ z_1 + z_2 \boldsymbol{i} ]_R = z_1$ and $[ z_1 + z_2 \boldsymbol{i} ]_I = z_2$. Thus, for $a,b \in \mathbb{C}$, we have
\[
\left[ \frac{\partial f_{CR}(a \cdot b)}{\partial b} \right]_R = \frac{\partial [f_{CR}(a \cdot b)]_R}{\partial [b]_R} = \frac{\partial [f_{CR}(a \cdot b)]_I}{\partial [b]_I}
\]
and
\[
\left[ \frac{\partial f_{CR}(a \cdot b)}{\partial b} \right]_I = \frac{\partial [f_{CR}(a \cdot b)]_I}{\partial [b]_R} = -\frac{\partial [f_{CR}(a \cdot b)]_R}{\partial [b]_I} .
\]

Finally, we introduce some notations throughout this work. For $\boldsymbol{w} \in \mathbb{C}^{n}$ and $\mathbf{W} \in \mathbb{C}^{n \times m}$, we denote by
\[
\| \boldsymbol{w} \|_2 \overset{\underset{\mathrm{def}}{}}{=} \left( \sum_{i=1}^{n}  |w_i|^2 \right)^{\frac{1}{2}} ~\!\text{and}~~ \| \mathbf{W} \|_2 \overset{\underset{\mathrm{def}}{}}{=} \left(\sum_{i=1}^{n} \sum_{j=1}^{m} |w_{ij}|^2\right)^{\frac{1}{2}} 
\]
the `entry-wise' vector norm and matrix norm, respectively. Let $[N]=\{1,2,\dots,N\}$ for integer $N>0$. Given a function $g(n)$, we denote by $h_1(n)=\Theta(g(n))$ if there exist positive constants $c_1, c_2$ and $n_0$ such that $c_1g(n) \leq h_1(n) \leq c_2g(n)$ for every $n \geq n_0$; we also denote by $h_2(n)=\mathcal{O}(g(n))$ if there exist positive constants $c$ and $n_0$ such that $h_2(n) \leq cg(n)$ for every $n \geq n_0$. 

We say that $f$ is a radial function if $f(\boldsymbol{x}) = f(\boldsymbol{x}')$ for $\|\boldsymbol{x}\| = \|\boldsymbol{x}'\|$. Let $\phi^2(\boldsymbol{x})$ be the density function of some probability measure $\mu$, which satisfies
\begin{equation} \label{eq:phi}
	\int_{\boldsymbol{x} \in \mathbb{R}^{2d}} \phi^2(\boldsymbol{x}) \dif \boldsymbol{x} = \int_{\boldsymbol{x} \in \mathcal{B}} 1 \dif \boldsymbol{x} = 1,
\end{equation}
where $\mathcal{B}$ is a unit ball. Then, for continuous functions $f,g$, we have the following equalities under Fourier transform
\begin{equation} \label{eq:fourier}
	\| \widehat{f\phi} - \widehat{g\phi} \|_{L_2{(\mu)}} = \| f\phi - g\phi \|_{L_2{(\mu)}} ,
\end{equation}
and 
\[
\widehat{f\phi} = \hat{f} * \hat{\phi}, \quad\text{for}\quad \text{the convolution operator} *.
\]

\section{Approximation} \label{sec:approximation}
We now present the first main theorem for complex-reaction networks as follows:
\begin{theorem} \label{thm:paras}
	For PReLU activation function, there exist a probability measure $\mu$ and a radial function $f:\mathbb{R}^{2d} \to \mathbb{R}$ such that
	\begin{itemize}
		\item[i)] for any $\delta>0$, there is a one-hidden-layer complex-reaction network $f_{CR}: \mathbb{C}^{d} \to \mathbb{R}$ with $O((d+1)(2d)^{3.75})$ parameters such that the followings hold:
		\[
		\mathbb{E}_{\boldsymbol{x} \sim \mu} (f_{CR}(\boldsymbol{x}) - f(\boldsymbol{x}))^2 < \delta ,
		\]
		\item[ii)] there exists a constant $\delta>0$ such that
		\[
		\mathbb{E}_{\boldsymbol{x} \sim \mu} \left( f_R(\boldsymbol{x}) - f(\boldsymbol{x}) \right)^2 \geq \delta ,
		\]
		for any one-hidden-layer real-valued network $f_R: \mathbb{R}^{2d} \to \mathbb{R}$ with exponential parameter number $O(C_1(2d+1)e^{C_1(2d)})$. Here, $C_1$ is an constant independent to $d$.
	\end{itemize}
\end{theorem}
This theorem shows that a kind of radial functions can be approximated by one-hidden-layer complex-reaction networks of polynomial parameters, whereas such functions cannot be approximated by real-valued networks with exponential ($O(C_1(2d+1)e^{C_1(2d)})$) parameters.

This proof idea can be summarized as follows. Provided the probability measure $\mu$ and corresponding density function $\phi$, defined as Eq.~\eqref{eq:phi}, it is observed that the composition $f\phi$ is radial as well as some radial function $f$. Further, we conjecture that $\widehat{f\phi}$ is also a radial function according to Eq.~\eqref{eq:fourier} with $g \equiv 0$. In contrast, the distribution of the composition $f_R\phi$ under Fourier transform, corresponding to a one-hidden-layer real-valued network $f_R$, is supported on a finite collection of lines $\{\boldsymbol{w}_i^{\top} \boldsymbol{x}\}$, i.e.,
\[
\textrm{Supp}(\widehat{f_R\phi}) \subseteq \left\{ \boldsymbol{x} ~\Big|~ \|\boldsymbol{x} -\boldsymbol{x}' \|_2 \leq1, \boldsymbol{x}' \in \bigcup_{i=1}^k \textrm{span}\{\boldsymbol{w}_i\} \right\} .
\]
Notice that the $\textrm{Supp}(\widehat{f_R\phi})$ is sparse in the Fourier space unless $k$ is an exponential. Thus, a one-hidden-layer real-valued network within polynomial parameters cannot achieve arbitrarily approximation for radial functions. On the other hand, the radial function is invariant to rotations, and is dependent on the input norm. It is observed that the zReLU function $\sigma_{cr}$ comprises the radius (i.e., norm) and phase (i.e., angle). Thus, there exist some linear connections (including rotation transformations) such that the combination of some complex-reaction neurons is invariant to rotations. In other words, complex-reaction networks can easily and well approximate some radial functions, see Lemma~\ref{lemma:complex_activations}.

Based on the ideas above, it is sufficient to provide an approximation guarantee between the complex-reaction network $f_{CR}$ and target function $f$, that is, (from Eq.~\eqref{eq:fourier})
\begin{multline*}
	\mathbb{E}_{\boldsymbol{x} \sim \mu} (f_{CR}(\boldsymbol{x}) - f(\boldsymbol{x}))^2 = \int  (f_{CR}(\boldsymbol{x}) - f(\boldsymbol{x}))^2 \phi^2(\boldsymbol{x}) \dif \boldsymbol{x} \\ = \| ( f_{CR}-f ) \phi \|_{L_2}^2(\mu) = \| \widehat{f\phi} - \widehat{g\phi} \|_{L_2{(\mu)}}^2 \leq \delta.
\end{multline*}
Motivated from \cite{eldan2016}, we consider the radial function
\begin{equation}\label{eq:function:f}
	f(\boldsymbol{x}) = \sum_{i=1}^N \epsilon_i f_i(\boldsymbol{x}) \quad \text{ with }\quad f_i(\boldsymbol{x}) = \boldsymbol{1}\{\|\boldsymbol{x}\|\in \Omega_i\} ,
\end{equation}
where $\epsilon_i \in \{-1, +1\}$, $N$ is a polynomial function of $d$, and $\Omega_i$'s are disjoint intervals of width $\mathcal{O}(1/N)$ on values in the range $\Theta(\sqrt{2d})$. Next, we formally begin our proof of Theorem~\ref{thm:paras} with some useful lemmas.
\begin{lemma} \label{lemma:real_activations}
	Let $f: [-R,R] \to \mathbb{R}$ be an $L$-Lipschitz function for constant $R>0$. For any $\delta>0$, $C_r \geq 1$, and $n_r \leq 2C_r LR / \delta$, there exists a real-valued network with one-hidden layer $f_R$ s.t.
	\[
	\sup_{x \in \mathbb{R}^{2d}} |f(x) - f_R(x)| \leq \delta .
	\]
	Specifically, $f_R: \mathbb{R} \to \mathbb{R}$ can be given by
	\[
	f_R(x) = \sum_{i=1}^{n_r} \alpha_i~ \sigma_r(\beta_i x - b_i) - a,
	\]
	for ReLU or general sigmoidal activation $\sigma_r$ and real-valued parameters $a, \{\alpha_i, \beta_i, b_i\}_{i=1}^{n_r}$.
\end{lemma}
Lemma~\ref{lemma:real_activations} shows that an one-dimensional $L$-Lipschitz function can be approximated by real-valued networks of one-hidden layer with general sigmoidal or ReLU activations. The detailed proof of Lemma~\ref{lemma:real_activations} is given in Appendix~\ref{app:real_activations}.

We now present a crucial lemma for complex-valued networks from Lemma~\ref{lemma:real_activations} as follows:
\begin{lemma} \label{lemma:complex_activations}
	Let $g: [r,R] \to \mathbb{R}$ be an $L$-Lipschitz function for $ r \leq R$. For any $\delta>0$, $C_{cr} \geq 1$, and $n_{cr} \leq 2C_{cr}dR^2L / (\sqrt{r}\delta)$, there exists a complex-reaction network with one-hidden layer $f_{CR}$ s.t.
	\[
	\sup_{\boldsymbol{x} \in \mathbb{C}^d} | g(\|\boldsymbol{x}\|) - f_{CR}(\boldsymbol{x}) | \leq \delta.
	\]
	Specifically, $f_{CR}: \mathbb{C}^{d} \to \mathbb{R}$ can be given by
	\[
	f_{CR}(\boldsymbol{x}) = \left[ \sum_{i=1}^{n_{cr}} v_i~ \sigma_{cr} (\boldsymbol{w}_i^T \boldsymbol{x} - b_i ) - a \right]_R ,
	\]
	for PReLU function $\sigma_{cr}$ and complex-valued parameters $a, \{\boldsymbol{w}_i, v_i, b_i\}_{i=1}^{n_{cr}}$.
\end{lemma}
\noindent\textbf{Proof:} Let $f': \mathbb{C}^d \to \mathbb{R}$ be a radial function with $f'(x) = | x |$. For any $\delta>0$ and $d \geq 1$, we have
\begin{equation} \label{eq:12}	
	\sup_{x \in \mathbb{C}} \Big| f'(x) - | \sigma_{cr}(w x - b) | \Big| \leq {\delta}/{2} .
\end{equation}
We further introduce a new function $g': \mathbb{R} \to \mathbb{R}$ as follows:
\[
g'(s) = \sum_{i=1}^{n'} \alpha_i' \sigma'(s) - a_i',
\]
where $\sigma$ is the ReLU function and $\alpha_i', a_i' \in \mathbb{R}$. For Lipschitz continuous function $\sqrt[r]{\cdot}$ and from Lemma~\ref{lemma:real_activations}, we have
\begin{equation} \label{eq:22}
	\sup_{s\in[r^k,R^k]} \left| g(\sqrt[k]{s}) - g'(s) \right| \leq {\delta}/{2},
\end{equation}
when $n' \leq C' L(R^k-r^k) / (\sqrt[k]{r}\delta)$ for some constant $C'>0$ and integer $k \geq 2$. Given complex-reaction network
\[
f_{CR}(\boldsymbol{x}) = \left[ \sum_{i=1}^{n_{cr}} v_i~ \sigma_{cr} (\boldsymbol{w}_i^{\top} \boldsymbol{x} - b_i ) - a \right]_R,
\]
we have
\begin{multline} \label{eq:32}
		\left| g'(s) - f_{CR}(\boldsymbol{x}) \right| = \left| g'(s) - f'(\boldsymbol{x}) \right| \\
		+ \left| f'(\boldsymbol{x}) - \left[ \sum_{i=1}^{n_{cr}} v_i~ \sigma_{cr} (\boldsymbol{w}_i^{\top} \boldsymbol{x} - b_i ) - a \right]_R \right| ,
\end{multline}
where
\[
f'(\boldsymbol{x}) = \sum_{i=1}^{n_{cr}'} v_i'~ \left| \sigma_{cr} (\boldsymbol{w}_i'^{\top} \boldsymbol{x} - b_i' ) \right| - a' .
\]
in which $\{\boldsymbol{w}_i',b_i'\}$ and $\{v_i'\}, a'$ denote another collection of complex-valued and real-valued parameters, respectively.

The first term of Eq.~\eqref{eq:32} can be bounded $\delta /4$ from Lemma~\ref{lemma:real_activations} for any $s \in [r^k, R^k]$. The second term is at most  $\delta /4$ when $n_{cr} \geq n_{cr}'$ from Eq.~\eqref{eq:12}. This follows that
\begin{equation} \label{eq:42}
	\left| g'(s) - f_{CR}(\boldsymbol{x}) \right| \leq  \delta / 2.
\end{equation}
Combining with Eqs.~\eqref{eq:22} and~\eqref{eq:42}, we have
\begin{multline*}
	\left| g(\|\boldsymbol{x}\|) - f_{CR}(\boldsymbol{x}) \right| \\ \leq  \left| g(\sqrt[k]{s}) - g'(s) \right|   +  \left| g'(s) - f_{CR}(\boldsymbol{x}) \right| \leq \delta ,
\end{multline*}
where $\boldsymbol{x}\in\mathbb{R}^{2d}\cong \mathbb{C}^d$ and $s \in [r^k, R^k]$. We finally obtain
\[
n_{cr} \leq 2C_{cr} (R^k-r^k) dL / (\sqrt[k]{r}\delta),
\]
provided $n_{cr} \leq 2dn'$ and $C' \leq C_{cr}$. We complete the proof by setting $k=2$ in the above upper bound.  $\hfill\square$
\begin{lemma} \label{lemma:complex_two}
	For $2d > C_2 > 0$, let $f:\mathbb{R}^{2d} \to \mathbb{R}$ is an $L$-Lipschitz radial function supported on the set
	\[
	\mathcal{S} = \{ \boldsymbol{x}: 0 < C_2\sqrt{2d} \leq \| \boldsymbol{x} \| \leq 2C_2\sqrt{2d} \}.
	\]
	For any $\delta >0$, there exists a complex-reaction network $f_{CR}$ of one-hidden layer with width at most $2C_{cr}(C_2)^{3/2}L(2d)^{7/4} / \delta$ such that
	\[
	\sup_{\boldsymbol{x}\in\mathbb{R}^{2d}\cong \mathbb{C}^d} | f(\boldsymbol{x}) - f_{CR}(\boldsymbol{x}) | < \delta.
	\]
\end{lemma}
Lemma~\ref{lemma:complex_two} shows that the radial functions can be approximated by complex-reaction networks with polynomial parameters, which is proved as follows.

\noindent\textbf{Proof:} Let $r = C_2\sqrt{2d}$, $R = 2C_2\sqrt{2d}$, and $d \geq 1$, then we have $r\geq 1$, which satisfies the condition of Lemma~\ref{lemma:complex_activations}. Invoke Lemma~\ref{lemma:complex_activations} to construct the complex-reaction networks and define $\delta' \leq \delta / 2d$. Then for any $L$-Lipschitz radial function $f:\mathbb{R}^{2d} \to \mathbb{R}$ supported on $\mathcal{S}$, we have
\[
\sup_{\boldsymbol{x}\in\mathbb{R}^{2d}\cong \mathbb{C}^d} | f(\boldsymbol{x}) - f_{CR}(\boldsymbol{x}) | \leq \delta',
\]
where the width of the hidden layer is bounded by
\[
n_{cr} \leq 2C_{cr}\frac{(C_2)^{3/2}dL}{\delta}(2d)^{3/4} \leq 2C_{cr}\frac{(C_2)^{3/2}L}{\delta}(2d)^{7/4}.
\]
This completes the proof. $\hfill\square$

\begin{lemma} \label{lemma:complex_one}
	Let $f(\boldsymbol{x}) = \sum_{i=1}^N \epsilon_i f_i(\boldsymbol{x})$ be defined by Eq.~\eqref{eq:function:f}. For any $\epsilon_i \in \{-1,+1\}$ $(i\in[N])$, there exists a Lipschitz function $g\colon \mathcal{S}\to [-1,+1]$ such that
	\[
	\int_{\mathbb{R}^{2d}} \left( g(\boldsymbol{x}) - f(\boldsymbol{x}) \right)^2 \phi^2(\boldsymbol{x}) ~d\boldsymbol{x} \leq  \frac{3}{(C_2)^2\sqrt{2d}}.
	\]
\end{lemma}
Lemma~\ref{lemma:complex_one} shows that any non-Lipschitz function $f(\boldsymbol{x})$ can be approximated and bounded by a Lipschitz function with density $\phi^2$, which is proved in Appendix~\ref{app:complex_one}. 

So far, the part (i) of Theorem~\ref{thm:approximation} can be proved as follows:
\begin{proposition} \label{prop:app_complex}
	Let $f$ be the radial function described by Eq.~\eqref{eq:function:f}.
	For $C_2, C_3>0$ with $d>C_2$, any $\delta>0$, and any choice of $\epsilon_i \in \{-1,+1\}$ $(i\in[N])$, there exists a complex-reaction network $f_{CR}$ of one-hidden layer with range in $[-2,+2]$ and width at most $C_3C_{cr}(2d)^{3.75}$, such that
	\[
	\left\| f(\boldsymbol{x}) - f_{CR}(\boldsymbol{x}) \right\|_{L_2(\mu)} \leq \frac{\sqrt{3}}{C_2 (2d)^{1/4}} + \delta.
	\]
\end{proposition}

\begin{lemma} \label{lemma:app_real}
	For positive constants $C_1, C_2, C_3, \rho, \alpha$ with $2d > C_2$ and $\alpha > C_2$, we define
	\[
	f(\boldsymbol{x}) = \sum_{i=1}^N \epsilon_i f_i(\boldsymbol{x}) \quad\text{and}\quad f_R(\boldsymbol{x}) = \sum_{i=1}^{n_r} \tilde{f}_i(\langle\boldsymbol{w}_i, \boldsymbol{x}\rangle),
	\]
	where $N \geq 4C_2 \alpha^{3/2}d^2$, $n_r \leq C_1 e^{2C_1d}$, and $\tilde{f}_i: \mathbb{R} \to \mathbb{R}$ are measurable functions with $|f_i(x)| \leq C_3(1+|x|^{\rho})$. For any $\delta >0$, there exists a group of $\epsilon_i \in \{-1,+1\}$ $(i\in[N])$ such that
	\[
	\| f(\boldsymbol{x}) - f_R(\boldsymbol{x}) \| \geq \delta / \alpha .
	\]
\end{lemma}
Lemma~\ref{lemma:app_real} shows that some radial function cannot be approximated by real-valued networks with exponential ($C_1e^{C_1(2d)}$) neurons, beyond which (ii) of Theorem~\ref{thm:approximation} holds. The detailed proof can be accessed in Appendix~\ref{app:app_real}.

\noindent\textbf{Proof of Theorem~\ref{thm:paras}:} Let $f(\boldsymbol{x}) = \sum_{i=1}^N \epsilon_i f_i(\boldsymbol{x})$ be defined by Eq.~\eqref{eq:function:f} and $N \geq 4C_2^{5/2}d^2$. According to Lemma~\ref{lemma:complex_one}, there exists a Lipschitz function $h$ with range $[-1,+1]$ such that
\[
\left\| h(\boldsymbol{x}) - f(\boldsymbol{x}) \right\|_{L_2(\mu)} \leq \frac{\sqrt{3}}{C_2 (2d)^{1/4}} .
\]
Based on Lemmas~\ref{lemma:complex_activations} and~\ref{lemma:complex_two}, any Lipschitz radial function supported on $\mathcal{S}$ can be approximated by a complex-reaction network $f_{CR}$ with one-hidden layer of width at most $C_3C_{cr}(2d)^{3.75}$, where $C_3$ is a constant relative to $C_2$ and $\delta$. This means that, 
\[
\sup_{\boldsymbol{x}\in\mathbb{R}^{2d}} | h(\boldsymbol{x}) - f_{CR}(\boldsymbol{x}) | \leq \delta .
\]
Thus, we have
\[
\| h - f_{CR} \|_{L_2(\mu)} \leq \delta.
\]
Hence, the range of $f_{CR}$ is in $[-1-\delta, +1+\delta] \subseteq [-2, +2]$. In summary, we have
\begin{multline*}
	\| f(\boldsymbol{x}) - f_{CR}(\boldsymbol{x}) \|_{L_2(\mu)} \leq \| f(\boldsymbol{x}) - h(\boldsymbol{x}) \|_{L_2(\mu)} \\ + \| h(\boldsymbol{x}) - f_{CR}(\boldsymbol{x}) \|_{L_2(\mu)} \leq \frac{\sqrt{3}}{C_2 (2d)^{1/4}} + \delta.
\end{multline*}
This implies that given constants $2d > C_2 >0$ and $C_3>0$, for any $\delta>0$ and $\epsilon_i \in \{-1,+1\}$ $(i\in[N])$, the target radial function $f$ can be approximated by a complex-reaction network $f_{CR}$ of one-hidden layer with range in $[-2,+2]$ and width at most $C_3C_{cr}(2d)^{3.75}$, that is,
\[
\| f_{CR} - f \|_{L_2(\mu)} \leq \frac{\sqrt{3}}{C_2 (2d)^{1/4}} + \delta < \delta_1.
\]
According to Lemmas~\ref{lemma:real_activations} and~\ref{lemma:app_real}, there are some groups of $\epsilon_i \in \{-1,+1\}$ $(i\in[N])$ such that
\[
\| f_R - f \|_{L_2(\mu)} \geq \delta_1 ,
\]
for any real-valued network $f_R$ of one-hidden layer with width at most $C_1C_r e^{2C_1d}$. The real-valued and complex-reaction networks have the number of parameters:
\[
\left\{\begin{aligned}
	N_r &= (2d) \times n_r  + n_r \leq C_1C_r(2d+1)e^{C_1(2d)}, \\
	N_{cr} &= 2 \times d \times n_{cr} + 2 \times n_{cr} \leq 2 C_3C_{cr}(d+1)(2d)^{3.75},
\end{aligned}\right.
\]
where $N_r$ and $N_{cr}$ indicate the parameter numbers of the real-valued and complex-reaction networks, respectively. This completes the proof.  $\hfill\square$

\section{Optimization Dynamics} \label{sec:OD}
This section studies the optimization dynamics of complex-reaction networks, and focuses on binary classification where $y \in \{-1, +1\}$ for simplicity. Let $\{\boldsymbol{x}_n,y_n\}_{n=1}^{N}$ be a training dataset, and denote by $\mathbf{X} = \{\boldsymbol{x}_n\}_{n=1}^{N}$. We employ the zReLU activation function, and use $[ f(\mathbf{W}; \boldsymbol{x}) ]_R$ to denote the complex-reaction network. We consider minimizing the following of empirical exponential loss:
\begin{equation} \label{eq:prob:minimizaiton}
	L( \mathbf{W};\mathbf{X} ) = \frac{1}{N} \sum_{n=1}^{N} \exp\left( y_n [f( \mathbf{W}; \boldsymbol{x}_n )]_R \right) ,
\end{equation}
where $\mathbf{W}=(\mathbf{W}^1,\mathbf{W}^2,\cdots,\mathbf{W}^L)$ for some integer $L>0$. We concern the solution sets of real-valued and complex-reaction networks by using standard gradient descents and weight normalization technique. Here, the weight normalization technique can be formalized by
\[
\mathbf{W}^l_j = \gamma^l_j~ \mathbf{V}^l_j, \quad \text{with} \quad \gamma^l_j \in \mathbb{R}^+ \quad \text{and} \quad \|\mathbf{V}^l_j\| =1,
\]
where $\mathbf{W}^l_j$ denotes the $i$-th row vector of the matrix $\mathbf{W}^l$ in the $l$-th layer.

We now present our second main result as follows:
\begin{theorem} \label{thm:minimas}
	For the minimization of Eq.~\eqref{eq:prob:minimizaiton}, we have
	\[
	\mathcal{G} \subseteq \mathcal{S}_{CR} \subsetneqq \mathcal{S}_R ,
	\]
	where $\mathcal{G}$ is the optima set of Eq.~\eqref{eq:prob:minimizaiton}, $\mathcal{S}_R$ and $\mathcal{S}_{CR}$ denote the critical point sets of real-valued and complex-reaction networks, respectively.
\end{theorem}
Theorem~\ref{thm:minimas} shows that the solution set of complex-reaction networks is a proper subset of that of real-valued networks, which may shed some insights on finding optimal solutions more easily for complex-reaction networks.

This proof idea can be summarized as follows: It forms a manifold for the parameters space of neural networks. Generally speaking, the complex manifold $\Omega_{CR}$ is a subset of the real manifold $\Omega_{R}$, since the coordinate transformation of complex manifold satisfies the holomorphic condition. On the other hand, it is essential to look for the critical points $\boldsymbol{\theta}$ with $\partial L(\boldsymbol{\theta}) / \partial \boldsymbol{\theta} =0$ when we solve an empirical minimization optimization using gradient descents. Therefore, the proof of Theorem~\ref{thm:minimas} can be converted into the problem of finding the critical points in $\Omega_{R}$ yet except $\Omega_{CR}$.

We construct a transformation to link the real and complex manifolds that correspond to the real-valued and complex-reaction networks, respectively. Finally, we find the desired points from the symplectic manifolds, which share certain characteristics with Riemannian geometry and complex geometry and link two geometric theories in some fields of mathematics.
\begin{definition} \label{def:phi}
	Define a linear mapping 
	$\phi: K^{p \times q} \to K^{p \times q}$,
	\[
	\begin{matrix}
		\phi(\Theta;\rho,i,j) = (\dots; &\underbrace{\rho\boldsymbol{\theta}_j+\boldsymbol{\theta}_i} &; \dots; &\underbrace{(1-\rho)\boldsymbol{\theta}_j}; \dots), \\
		& i & & j &
	\end{matrix}
	\]
	where $K^{p \times q}$ is compact in $\mathbb{R}^{p \times q}$ or $\mathbb{C}^{p \times q}$, $\Theta = ( \boldsymbol{\theta}_1, \dots, \boldsymbol{\theta}_q) \in K^{p \times q}$, $\rho \in$ $\mathbb{R}$ or $\mathbb{C}$, and $i,j \in [q]$.
\end{definition}
\begin{lemma} \label{lemma:complex_affine}
	For any $i,j \in [q]$, $\phi(\Theta)$ consists of some straight lines in an 2-dimensional affine space.
\end{lemma}
Lemma~\ref{lemma:complex_affine} shows that the linear mapping $\phi$ leads to an affine space defined by $\boldsymbol{\theta}_j$, which is proved by Appendix~\ref{app:complex_affine}.
\begin{definition}
	A mapping is said to be \textbf{analytic} if it is continuous and expandable in a power series around any points.
\end{definition}
\begin{definition} \label{def:T}
	Let $f$ be a function expressed by a neural network with parameter space $K$. An analytic mapping $\mathcal{T} \in \mathcal{C}^1(K)$ is said to be an \textbf{equioutput transformation}, if $f(\mathcal{T}(\mathbf{W}); \mathbf{X}) = f(\mathbf{W}; \mathbf{X})$ for any collection of parameters $\mathbf{V} \in K$.
\end{definition}
\begin{lemma} \label{lemma:composition}
	For any equioutput transformation $\mathcal{T} \in \mathcal{C}^1(K)$, there is a collection of finite mappings~$\{\phi\}$, such that $\mathcal{T}$ is a composition of $\phi$'s.
\end{lemma}
\begin{lemma} \label{lemma:complex_group}
	All equioutput transformations, generated from Lemma~\ref{lemma:composition}, constitute a multiplicative group $\mathbb{G}$, which is isomorphic to a direct product of Weyl groups.
\end{lemma}
A straightforward combination of Lemmas~\ref{lemma:composition} and~\ref{lemma:complex_group} shows that the equioutput transformation is composited of finite linear mapping $\phi$, and for any $i,j$, the equioutput transformations constitute an algebraic group, isomorphic to a direct product of Weyl groups~\cite{warner1983}. The detailed proof of Lemma~\ref{lemma:complex_group} are present as follows.

\noindent\textbf{Proof:} Write the parameter space as $\Theta^1 \times \dots \times \Theta^L$, where $\Theta^l$ denotes the subspace concerning the $l$-th layer. Let $\mathbb{G}_l$ denotes the set of linear mapping $\phi$ upon $\Theta^l$. According to Lemmas~\ref{lemma:composition} and~\ref{lemma:complex_affine}, $\mathbb{G}_l$ forms a cube symmetry group, that is, isomorphic to the Weyl group~\cite{warner1983,weyl1946}. So the equioutput transformation, i.e., group action upon each hidden layer (except the case $l = L$) can be regarded as the direct operation of $\mathbb{G}_l$ on the corresponding subspace $\Theta^1$ and as indirect but isomorphic operation led by some sequence $\{\phi\}$. According to Lemma~\ref{lemma:composition} and the fact that the hidden layers only have symmetry groups associated with themselves, each hidden layer contributes exactly one cube symmetry group to the overall group action. In other words, $\mathbb{G}_l$ 1-1 corresponds to $\Theta^l$. Thus, group $\mathbb{G}$ is isomorphic to the direct product of these groups
\begin{equation} \label{eq:potential}
	\mathbb{G} \cong \mathbb{G}_1 \times \dots \times \mathbb{G}_{L-1},
\end{equation}
since the actions of the individual groups operating on different layers commute. Next, we are going to bound the order of $\mathbb{G}$, denoted as $|\mathbb{G}|_{\#}$. Based on Eq.~\eqref{eq:potential}, we have
\[
|\mathbb{G}|_{\#} = \prod_{l=1}^{L-1} |\mathbb{G}_l|_{\#}.
\]
Suppose that the $l$-th layer has $n_l$ neurons, then there are $n_l!$ different pairs $(i,j)$ in this layer. Further, if the equation $ L(\Theta^l,\alpha_i,\alpha_j ) = L(\phi(\Theta^l,\rho,i,j),\tilde{\boldsymbol{\alpha}}_i,\tilde{\boldsymbol{\alpha}}_j) $ in the proof of Lemma~\ref{lemma:composition}, led by the minimum generator, has finite solution, then from the Weyl group theory, the order of the group $|\mathbb{G}_l|_{\#}$ is bounded by $n_l! 2^{n_l}$. Otherwise, $\mathbb{G}$ is an infinite group. This completes the proof. $\hfill\square$

Next, we provide two crucial propositions about the dynamical systems led by complex-reaction and real-valued networks with standard gradient descents, respectively.
\begin{proposition} \label{prop:gradients}
	For the minimization of Eq.~\eqref{eq:prob:minimizaiton} using complex-reaction networks, we have the following dynamical systems
	\begin{equation} \label{eq:complex_gradients}
		\left\{ \begin{aligned}
			\frac{d \gamma^l_j}{d t} &= \frac{\eta}{\gamma^l_j} \frac{1}{N} \sum_{n=1}^N \exp\left( y_n [f( \mathbf{W}; \boldsymbol{x}_n )]_R \right) y_n [f( \mathbf{V}; \boldsymbol{x}_n )]_R ,\\
			\frac{d \mathbf{V}^l_j}{d t} &= \frac{\eta}{\left(\gamma^l_j\right)^2} \frac{1}{N} \sum_{n=1}^N \exp\left( y_n [f( \mathbf{W}; \boldsymbol{x}_n )]_R \right) y_n \Delta_j ,\\
		\end{aligned} \right.
	\end{equation}
	where $\eta$ is a strictly positive constant relative to $\gamma^l_j$, and
	\begin{multline*}
		\Delta_j = \left( \frac{\partial [f( \mathbf{V}; \boldsymbol{x}_n )]_R}{\partial \left[ \mathbf{V}_j^l \right]_R } - [\mathbf{V}_j^l]_R [f( \mathbf{V}; \boldsymbol{x}_n )]_R \right) \\+ \left(  \frac{\partial [f( \mathbf{V}; \boldsymbol{x}_n )]_I}{\partial \left[ \mathbf{V}_j^l \right]_I } - [\mathbf{V}_j^l]_I [f( \mathbf{V}; \boldsymbol{x}_n )]_R \right) \boldsymbol{i} .
	\end{multline*}
\end{proposition}
\begin{proposition} \label{lemma:gradients_r}
	Let $f_R:\mathbb{R}^{2d} \to \{-1,+1\}$ be a real-valued neural network with ReLU activation and weight normalization $\mathbf{P}^l_j = \gamma_j^l \mathbf{Q}^l_j$ where $\| \mathbf{Q}^l_j \|=1$. the gradient descent procedure for minimizing the exponential loss coincides with the following dynamical systems
	\begin{equation} \label{eq:real_gradients}
		\left\{ \begin{aligned}
			\frac{d \gamma^l_j}{d t} &= \frac{\eta}{\gamma^l_j} \frac{1}{N} \sum_{n=1}^N \exp\left( -y_n f_R( \mathbf{P}; \boldsymbol{x}_n ) \right) y_n f_R( \mathbf{Q}; \boldsymbol{x}_n ) ,\\
			\frac{d \mathbf{Q}^l_j}{d t} &= \frac{\eta}{\left(\gamma^l_j\right)^2} \frac{1}{N} \sum_{n=1}^N \exp\left( -y_n f_R( \mathbf{P}; \boldsymbol{x}_n ) \right) y_n \boldsymbol{\phi}_j^l ,\\
		\end{aligned} \right.
	\end{equation}
	where $\eta$ is a strictly positive constant relative to $\gamma^l_j$ and
	\[
	\boldsymbol{\phi}_j^l = \frac{\partial f_R( \mathbf{Q}; \boldsymbol{x}_n )}{\partial \mathbf{Q}_j^l} - \mathbf{Q}_j^l  f_R( \mathbf{Q}; \boldsymbol{x}_n ) .
	\]
\end{proposition}
Propositions~\ref{prop:gradients} and~\ref{lemma:gradients_r} hold from Lemmas~\ref{lemma:matrix} and~\ref{lemma:gamma} as follows:
\begin{lemma}[Normalization Matrix] \label{lemma:matrix}
	Let $\boldsymbol{w}, \boldsymbol{v} \in \mathbb{R}^{1\times n}$ with $\boldsymbol{w} = \gamma \boldsymbol{v}$ and $\| \boldsymbol{v} \|=1$, and $\mathbf{S} = \mathbf{I}_{n\times n} - \boldsymbol{v}^T\boldsymbol{v}$, then we have
	\[
	(1)~ \mathbf{S} = \mathbf{I}_{n\times n} - \frac{\boldsymbol{w}^T \boldsymbol{w}}{\|\boldsymbol{w}\|_2^2}; \quad (2)~ \frac{\partial \boldsymbol{v}}{\partial \boldsymbol{w}} = \frac{\mathbf{S}}{\gamma};
	\]
	and
	\[
	\quad (3)~ \mathbf{S} \boldsymbol{w}^T = \mathbf{S} \boldsymbol{v}^T = \boldsymbol{0}; \quad (4)~ \mathbf{S}^2 = \mathbf{S}.
	\]
\end{lemma}
\begin{lemma}[Weight Norms] \label{lemma:gamma} 
	During the gradient descent procedure, the change rate of $\| \gamma^l_j \|$ (i.e., weight norms) is the same for each layer.
\end{lemma}
Lemmas~\ref{lemma:matrix} and~\ref{lemma:gamma} stand up for both complex-reaction and real-valued networks. Notice that Lemma~\ref{lemma:gamma} shows that $(\gamma^l_j)^2 = \| \mathbf{W}^l_j \|^2$ grows at a rate independent of the row $j$ and layer $l$. Thereby, using gradient descent to solve the optimization problem, there is no difference in the change rate of connection weights layer by layer. This result also holds for real-valued networks. The detailed proofs of Lemmas~\ref{lemma:matrix} and~\ref{lemma:gamma} are presented by Appendix~\ref{app:matrix} and~\ref{app:gamma}, respectively.

Based on the aforementioned results, we now present the crucial lemmas for proving Theorem~\ref{thm:minimas} as follows:

\begin{lemma} \label{lemma:complex_core} Let $\mathbf{V}$ be the complex-valued parameters of a complex-reaction network. If $\mathbf{V}$ is a critical point of $L(\mathbf{V})$, then we have
	\begin{itemize}
		\item[i)] $ \partial L(\mathcal{T}(\mathbf{V})) / \partial \mathbf{V} = 0$ for any equioutput transformation $\mathcal{T}$ with $\rho\in\mathbb{R}$;
		\item[ii)] $\partial L(\mathcal{T}(\mathbf{V})) / \partial \mathbf{V} \neq 0$ for any equioutput transformation $\mathcal{T}$ with $\rho\in\mathbb{C}$.
	\end{itemize}
\end{lemma}
\noindent\textbf{Proof:} From Lemma~\ref{lemma:composition}, any transformation is composited of $\phi$'s. Thus, it suffices to prove that the conclusions above hold upon the minor equioutput transformation from the basic theorem of algebra. Here, we consider building the following minor equioutput transformation.

Let $\Theta \in \mathbb{C}^{p \times q}$ and $\Lambda = (\alpha_{sk}) \in \mathbb{C}^{q \times r}$ denote the connection weights of adjacent layers in a complex-reaction network, respectively. For any $i,j\in[q]$, $s \in [r]$, and $\rho,\rho'\in \mathbb{C}$, we abbreviate $\phi(\Theta;\rho,i,j)$, $\rho\boldsymbol{\theta}_j + \boldsymbol{\theta}_i$, $(1-\rho)\boldsymbol{\theta}_j$, $\rho'\alpha_{sj} + \boldsymbol{\theta}_{si}$, and $(1-\rho')\alpha_{sj}$ as $\tilde{\Theta}$, $\tilde{\boldsymbol{\theta}}_i$, $\tilde{\boldsymbol{\theta}}_j$, $\tilde{\alpha}_{si}$, and $\tilde{\alpha}_{sj}$, respectively. Hence, the output of the original network
\[
h_s(\boldsymbol{z}) = \boldsymbol{\alpha}_s \sigma_{cr}\left( \Theta \boldsymbol{z} \right) ,
\]
where $\boldsymbol{\alpha}_s$ denotes the $s$-th row vector of $\Lambda$ and $\boldsymbol{z} \in \mathbb{C}^{p \times 1}$ is the input. Compositing the linear mapping $\phi$ with $\Theta$, we have the output of the transformed network
\[
\begin{aligned}
	\tilde{h}_s(\boldsymbol{z}) =&~ \sum_{k\neq i,j}^{q} \alpha_{sk} \sigma_{cr}\left( \boldsymbol{\theta}_k^{\top} \boldsymbol{z} \right) + \tilde{\alpha}_{si}\sigma_{cr}\left( (\rho\boldsymbol{\theta}_j + \boldsymbol{\theta}_i)^{\top} \boldsymbol{z} \right) \\&+ \tilde{\alpha}_{sj} \sigma_{cr}\left( (1-\rho)\boldsymbol{\theta}_j^{\top} \boldsymbol{z} \right) . 
\end{aligned}
\]
Let $\tilde{h}_s(\boldsymbol{z})=h_s(\boldsymbol{z})$. The equation has at least one solution since the degree of freedom (i.e., 4) of this equation is greater than the number of equations (i.e., 2). It implies that for any $\rho$, there exists a linear mapping $\phi'$ with $\rho'$ acts upon the connection weights $\Lambda$, such that
\[
L(\Theta,\Lambda) = L(\tilde{\Theta},\phi'(\Lambda)) = L(\phi(\Theta),\phi'(\Lambda)) ,
\]
where $L(\Theta,\Lambda)$ is a short notation of loss function described in Eq.~\eqref{eq:prob:minimizaiton}. From Lemma~\ref{lemma:complex_group}, the stacking of $\phi$ and $\phi'$ constitutes the minimum generator of $\mathbb{G}$. Hence, the composition of $\phi$ and $\phi'$ is the desired minor equioutput transformation. 

Based on Eq.~\eqref{eq:complex_gradients} and $ \partial L(\Theta,\Lambda)/\partial \boldsymbol{\theta}_k = 0 $ ($k \in [q]$), it stands for the original networks
\[
\frac{1}{N} \sum_{n=1}^N y_n \exp(- y_n [f( \mathbf{W}; \boldsymbol{x}_n )]_R) \Delta(\boldsymbol{\theta_k};\boldsymbol{x}_n) = 0 ,
\]
where
\begin{multline*}
	\Delta(\boldsymbol{\theta_k};\boldsymbol{x}_n) = \left( \frac{\partial [f( \Theta; \boldsymbol{x}_n )]_R}{\partial \left[ \boldsymbol{\theta}_k \right]_R } - [\boldsymbol{\theta}_k]_R [f( \Theta; \boldsymbol{x}_n )]_R \right) \\ + \left(  \frac{\partial [f( \Theta; \boldsymbol{x}_n )]_I}{\partial \left[ \boldsymbol{\theta}_k \right]_I } - [\boldsymbol{\theta}_k]_I [f( \Theta; \boldsymbol{x}_n )]_R \right) \boldsymbol{i} .
\end{multline*}
For the transformed networks, we consider the following cases. (a) For $k \neq i,j$, we have
\[
\frac{\partial L(\tilde{\Theta},\phi'(\Lambda))}{\partial \tilde{\boldsymbol{\theta}}_k} = \frac{\partial L(\Theta,\Lambda)}{\partial \boldsymbol{\theta}_k} = 0 .
\]
(b) For $i$, we have
\[
\frac{\partial L(\tilde{\Theta},\phi'(\Lambda))}{\partial \tilde{\boldsymbol{\theta}}_i} \propto \frac{1}{N} \sum_{n=1}^N \frac{1}{r} \sum_{s=1}^r y_n L(\Theta,\Lambda) ~\zeta_i ,
\]
where $\tilde{\boldsymbol{\alpha}}_i$ denotes the $i$-th row vector of $\phi'(\Lambda)$, and
\[
\begin{aligned}
	\zeta_i &= \frac{\partial [f( \Theta; \boldsymbol{z}_s, \boldsymbol{x}_n )]_R}{\partial \boldsymbol{h} } \tilde{\boldsymbol{\alpha}}_i \frac{\partial \sigma_{cr}( \Theta \boldsymbol{z} )}{\partial \boldsymbol{z}} \boldsymbol{z} \\&+ \frac{\partial [f( \Theta; \boldsymbol{z}_s, \boldsymbol{x}_n )]_R}{\partial \bar{\boldsymbol{h}} } \overline{\tilde{\boldsymbol{\alpha}}}_i \frac{\partial \overline{\sigma_{cr}( \Theta \boldsymbol{z} )}}{\partial \boldsymbol{z}} \boldsymbol{z} .
\end{aligned}
\]
Thus, we have
\[
\frac{\partial L(\tilde{\Theta},\phi')}{\partial \tilde{\boldsymbol{\theta}}_i} \left\{ \begin{aligned}
	= 0, &~~\text{if}~~ \rho \in \mathbb{R}; \\
	\neq0, &~~\text{if}~~ \rho \in \mathbb{C} ~~\text{and}~~ [\rho]_I \neq 0 .
\end{aligned} \right.
\]
(c) Similarly, for $j$, it holds
\[
\frac{\partial L(\tilde{\Theta},\phi')}{\partial \tilde{\boldsymbol{\theta}}_j} \left\{ \begin{aligned}
	= 0, &~~\text{if}~~ \rho \in \mathbb{R}, \\
	\neq0, &~~\text{if}~~ \rho \in \mathbb{C} ~~\text{and}~~ [\rho]_I \neq 0 .
\end{aligned} \right.
\]
Lemma~\ref{lemma:complex_core} holds as desired when $\Theta = \mathbf{V}^l$  ($l \neq L$). $\hfill\square$

From Lemma~\ref{lemma:complex_core}, it is easy to obtain the following lemma.
\begin{lemma} \label{lemma:real_core}
	Let $\mathbf{Q}$ be the real-valued parameters of a real-valued neural network. If $\mathbf{Q}$ is a critical point of $L(\mathbf{Q})$, then for any equioutput transformation $\mathcal{T}$ with $\rho\in\mathbb{R}$, $\phi(\mathcal{T}(\mathbf{Q}))$ are critical points.
\end{lemma}
\begin{lemma} \label{lemma:manifold}
	For any fully-connected feed-forward real-valued neural network with parameter space $\Omega_{R}$, there exists a complex-reaction network with parameter space $\Omega_{CR}$ such that $\Omega_{CR} \subseteq \Omega_{R}$.
\end{lemma}
This proposition shows that, for any real-valued neural network, we can construct a complex-reaction network such that i) both networks have the same depth, and ii) both networks have the same number of parameters for each layer. The detailed proof is given by Appendix~\ref{app:manifold}.

\noindent\textbf{Proof of Theorem~\ref{thm:minimas}:} Let $\mathbb{G}_R$ and $\mathbb{G}_{CR}$ denote the groups of the equioutput transformations with real-valued and complex-valued $\rho$ (Definition~\ref{def:phi}), respectively. Let $\mathcal{G}$ be the optima set, $\mathcal{S}_R$ and $\mathcal{S}_{CR}$ denote the critical point sets of the real-valued and complex-reaction networks, respectively. It suffices to consider the real-valued and complex-reaction networks with the same parameters, generated from Proposition~\ref{lemma:manifold}, and this follows $\Omega_{CR} \subseteq \Omega_{R}$.

From Lemma~\ref{lemma:complex_affine}, we divide the $4 \times 4$ matrices in Eq.~\eqref{eq:rho_r} and~\eqref{eq:rho_c} into the block formations by $2\times2$, and find that two block matrices relative to $\boldsymbol{\theta_j}$ are anti-symmetric for $\rho \in \mathbb{C}$ (see Eq.~\eqref{eq:rho_r}), and thus $\phi$ is a linear mapping from complex manifold to complex manifold. For $\rho \in \mathbb{R}$, two block matrices relative to $\boldsymbol{\theta_j}$ are diagonal (see Eq.~\eqref{eq:rho_c}), which implies that $\phi(\boldsymbol{\theta_j})$ is not on the complex manifold. In other words, the equioutput transformation with real-valued $\rho$ projects a critical point on complex manifold onto the real manifold, specially \emph{almost-complex manifold}~\cite{newlander1957,wells1980}. The transformed points are still critical points from Lemma~\ref{lemma:complex_core}(i), whereas any critical point in complex manifold after any equioutput transformation with complex-valued $\rho$ cannot derive new critical points from Lemma~\ref{lemma:complex_core}(ii). On the other hand, Lemma~\ref{lemma:real_core} shows that a critical point on real manifold after any equioutput transformation still dwells on the real manifold and derives new critical points.

In summary, the critical point set of real-valued networks are closed to the equioutput transformation with real-valued $\rho$, where this property does not hold for that of complex-reaction networks. Therefore, we have
\begin{equation} \label{eq:group1}
	\left\{\begin{aligned}
		&\mathcal{G} \subseteq \mathbb{G}_R \circ \mathcal{S}_{CR} \subseteq \mathbb{G}_R \circ \mathcal{S}_R = \mathcal{S}_R \\ &\mathcal{G} \subseteq \mathcal{S}_{CR} \subsetneqq \mathbb{G}_{CR} \circ \mathcal{S}_{CR}.
	\end{aligned}
	\right.
\end{equation}
From Lemma~\ref{lemma:complex_group}, both groups $\mathbb{G}_R$ and $\mathbb{G}_{CR}$ are isomorphic to a direct product of Weyl groups. Thus, $\mathbb{G}_R$ is isomorphic to $\mathbb{G}_{CR}$, i.e., $ \mathbb{G}_R \cong \mathbb{G}_{CR} $. This complete the proof.  $\hfill\square$

\section{Conclusions}  \label{sec:conclusion}
This work presents some theoretical understanding of complex-valued neural networks beyond real-valued ones. We introduce the complex-reaction network with fully-connected feed-forward architecture and provide the universal approximation property for complex-reaction networks. We show that a class of radial functions can be approximated by a complex-reaction network using the polynomial number of parameters, yet cannot be approximated by real-valued networks with exponential parameters. We also prove that for practical optimization problems, the critical point set of complex-reaction networks is a proper subset of that of real-valued networks, which may shed some insights on finding optimal solutions more easily for complex-reaction networks. In the future, it is interesting to explore other advantages of complex-valued neural networks beyond real-valued ones, such as from the perspective of generalization.

\bibliography{reference}
\bibliographystyle{plain}

\onecolumn
\appendix
\begin{center}
	{\Large\textbf{Supplementary Materials of Towards Understanding Theoretical Advantages of Complex-Reaction Networks (Appendix)}}
\end{center}

\section{Complete Proofs for Theorem~\ref{thm:paras}}
We provided the detailed proofs for Theorems~\ref{thm:paras}.

\subsection{Constructions in Theorem~\ref{thm:paras}} \label{app:approximation}
Here, we first review the candidate radial function in~\cite{eldan2016}. Define
\[
f(\boldsymbol{x}) = \sum_{i=1}^N \epsilon_i f_i(\boldsymbol{x}) ,
\]
where $\epsilon_i \in \{-1, +1\}$, $N$ is a polynomial function of $d$,
\[
f_i(\boldsymbol{x}) = \left\{ \begin{aligned}
	\boldsymbol{1}, &~~\text{if}~~B_i = 1, \\
	\boldsymbol{0}, &~~\text{if}~~B_i = 0,
\end{aligned}\right.
\]
for $B_i$'s are binary indicators, and
\[
\Omega_i = \left[ \left( 1+ \frac{i-1}{N} \right) C_2 \sqrt{2d}, \left( 1+\frac{i}{N} \right) C_2 \sqrt{2d} \right] ,~~ i = 1,\dots,N.
\]

Next, the concerned density function $\phi$ is the Fourier transform of the indicator of a unit-volume Euclidean ball, that is,
\[
\int_{\mathbb{R}^{2d}} \phi^2(\boldsymbol{x}) \dif \boldsymbol{x} = \int_{\mathcal{B}_d} \boldsymbol{1} \dif \boldsymbol{\omega} = 1.
\]
Thus, we have
\[
\phi(\boldsymbol{x}) = \int_{\mathcal{B}_d} 1 \cdot \exp\left( -2\pi \boldsymbol{i} \boldsymbol{x}^{\top} \boldsymbol{\omega} \right) \dif \boldsymbol{\omega} \quad \text{with} \quad \mathcal{B}_d = \left\{ \boldsymbol{\omega} ~\Big|~ \| \boldsymbol{\omega} \|_2 \leq \left( \int_{\mathcal{B}} \boldsymbol{1} \dif \boldsymbol{\omega} \right)^{-1} \right\} .
\]
The proof of Theorem~\ref{thm:approximation} consists of the following two parts:

\noindent\textbf{(i) of Theorem~\ref{thm:approximation}.} Let $f$ be the radial function described by Eq.~\eqref{eq:function:f}. For $C_2, C_3>0$ with $d>C_2$, any $\delta>0$, and any choice of $\epsilon_i \in \{-1,+1\}$ $(i\in[N])$, there exists a complex-reaction network $f_{CR}$ of one-hidden layer with range in $[-2,+2]$ and width at most $C_3C_{cr}(2d)^{3.75}$, such that
\[
\left\| f(\boldsymbol{x}) - f_{CR}(\boldsymbol{x}) \right\|_{L_2(\mu)} \leq \frac{\sqrt{3}}{C_2 (2d)^{1/4}} + \delta.
\]

\noindent\textbf{(ii) of Theorem~\ref{thm:approximation}.} For positive constants $C_1, C_2, C_3, \rho, \alpha$ with $2d > C_2$ and $\alpha > C_2$, we define
\[
f(\boldsymbol{x}) = \sum_{i=1}^N \epsilon_i f_i(\boldsymbol{x}) \quad\text{and}\quad f_R(\boldsymbol{x}) = \sum_{i=1}^{n_r} \tilde{f}_i(\langle\boldsymbol{w}_i, \boldsymbol{x}\rangle),
\]
where $N \geq 4C_2 \alpha^{3/2}d^2$, $n_r \leq C_1 e^{2C_1d}$, and $\tilde{f}_i: \mathbb{R} \to \mathbb{R}$ are measurable functions with $|f_i(x)| \leq C_3(1+|x|^{\rho})$. For any $\delta >0$, there exists a group of $\epsilon_i \in \{-1,+1\}$ $(i\in[N])$ such that
\[
\| f(\boldsymbol{x}) - f_R(\boldsymbol{x}) \| \geq \delta / \alpha .
\]

\subsection{Proof of Lemma~\ref{lemma:real_activations}} \label{app:real_activations}
From the universal approximation theorems~\citep{cybenko1989,leshno1993}, the shallow real-valued neural networks with general sigmoidal and ReLU activation functions have the universal approximation properties for any continuous functions. Here, we provide the tight bounds for the degree of approximation.

For general sigmoidal activation, \citet{debao1993} has proved that a $L$-Lipschitz function $f$ can be approximated by real-valued neural networks of one-hidden layer with $\delta \geq L\omega(f; n_r^{-1})$ .

Next, we discuss the ReLU activation. Given $2d=1$, we have $R \geq \delta / (2L)$; Otherwise, Lemma~\ref{lemma:real_activations} is trivially satisfied once we force the real-valued network $f_R \equiv 0$. In the case of $R \geq \delta / (2L)$, let $n_0 = \lceil \delta / (2RL) \rceil$. For any $L$-Lipschitz function $f:[-R,R] \to \mathbb{R}$, we have 
\[
|g_i(\beta) - f(\beta)| \leq \delta ,
\]
where
\[
g_i(x) = g_i(-R) + \frac{g_i(\beta+\delta/(2L)) - g_i(\beta-\delta/(2L))}{\delta/L} \sigma_r(x-\beta) \quad\text{with}\quad \beta = \delta/L \quad\text{for}\quad i\in[n_0].
\]
Thus, then we have 
\[
\sup_{x \in \mathbb{R}} \big| f(x) - g(x) \big| \leq \delta,
\]
where 
\[
g(x) = g(-R) + \sum_{i}^{n_0} \frac{g_i(\beta_i+\delta/(2L)) - g_i(\beta_i-\delta/(2L))}{\delta/L} \sigma_r(x-\beta_i)  ,
\]
and
\[
|g_i(\beta_i) - f(\beta_i)| \leq \delta , \quad\text{for}\quad \beta_i = i \delta/L \quad\text{and}\quad i\in[n_0] .
\]
Provided a real-valued neural network of one-hidden layer with ReLU activation
\[
f_R(x) = \sum_{i=1}^{n_r} \alpha_i~ \sigma_r(\beta_i x - b_i) - a,
\]
we employ that
\[
a = -g(-R), \quad \alpha_i = \frac{g_i(\beta_i+\delta/(2L)) - g_i(\beta_i-\delta/(2L))}{\delta/L}, \quad \beta_i = i \delta/L, \quad\text{and}\quad n_r \leq n_0 \leq \frac{\delta}{2RL} ,
\]
and then, Lemma~\ref{lemma:real_activations} holds as desired. $\hfill\square$

\subsection{Proof of Lemma~\ref{lemma:complex_one}} \label{app:complex_one}
Define a branch function
\[
g_i(\boldsymbol{x}) = \left\{ \begin{aligned}
	\max\{ \boldsymbol{1}, ND_i \}, &~~\text{if}~~B_i = 1, \\
	\boldsymbol{0}~~~~~~~~~~~, &~~\text{if}~~B_i = 0,
\end{aligned}\right.
\]
with
\[
D_i = \min\left\{ \left| \|\boldsymbol{x}\| - \left( 1+ \frac{i-1}{N} \right) C_2 \sqrt{2d}  \right|, \left| \|\boldsymbol{x}\| - \left( 1+ \frac{i}{N} \right) C_2 \sqrt{2d}  \right| \right\} .
\]
Let
\[
g(\boldsymbol{x})  = \sum_{i=1}^N \epsilon_i g_i(\boldsymbol{x}).
\]
Since $B_i = 1$ and $\Omega_i$'s are disjoint intervals, $g_i(\boldsymbol{x})$ is an $N$-Lipschitz function. Thus, $g$ is also an $N$-Lipschitz function. So we have
\begin{multline*}
	\int_{\mathbb{R}^{2d}} \left( g(\boldsymbol{x}) - \sum_{i=1}^N \epsilon_i f_i(\boldsymbol{x}) \right)^2 \phi^2(\boldsymbol{x}) ~d\boldsymbol{x}  = \int_{\mathbb{R}^{2d}}  \sum_{i=1}^N \epsilon_i^2 \left( g_i(\boldsymbol{x}) - f_i(\boldsymbol{x}) \right)^2 \phi^2(\boldsymbol{x}) ~d\boldsymbol{x} \\
	= \sum_{i=1}^N \int_{\mathbb{R}^{2d}} \left( g_i(\boldsymbol{x}) - f_i(\boldsymbol{x}) \right)^2 \phi^2(\boldsymbol{x}) ~d\boldsymbol{x}\leq (3/(C_2)^2\sqrt{2d}),
\end{multline*}
where the last inequality holds from \cite[Lemma 22]{eldan2016}. This completes the proof. $\hfill\square$

\subsection{Proof of Lemma~\ref{lemma:app_real}} \label{app:app_real}
We first list some useful technical conclusions.
\begin{proposition} \label{prop:2}
	Let $f_R(\boldsymbol{x}) = \sum_{i=1}^k \tilde{f}_i(\langle\boldsymbol{w}_i, \boldsymbol{x}\rangle)$, where $\tilde{f}_i: \mathbb{R} \to \mathbb{R}$ are measurable functions satisfying $|f_i(x)| \leq C_3(1+|x|^{\rho})$ and $\rho$ is an integer satisfying $\rho \leq C_1 e^{2C_1d}$. If $f_R \phi \in L_2$, then
	\[
	\text{Supp}(\widehat{f_R\phi}) \subseteq \bigcup_{i=1}^k \left( \text{span}\{\boldsymbol{w}_i\} + \mathcal{B} \right),
	\]
	where $\mathcal{B}$ is a unit ball. Furthermore, there exists a pair of functions $(p,q)$ that satisfies
	\begin{itemize}
		\item $p \in \text{Supp}(\widehat{f_R\phi})$;
		\item $q$ is radial and $\int_{B}q(\boldsymbol{x})^2d\boldsymbol{x}\leq1-\delta$ for some $\delta\in[0,1]$;
		\item $\|p\|_{L_2} = \|q\|_{L_2} =1$. Then
		\[
		1- \langle p,q\rangle_{L_2} \geq \delta/2 - k e^{-2Cd}.
		\]
	\end{itemize}
\end{proposition}
This proposition is proved by~\cite{eldan2016}.
\begin{proposition} \label{prop:1}
	Provided $\|p\|_{L_2} = \|q\|_{L_2} =1$, for any real-valued scalars $a, b>0$, we have
	\[
	\|ap - bq\|_{L_2} \geq \frac{b}{2} \|p - q\|_{L_2}.
	\]
\end{proposition}
\begin{proposition} \label{prop:3}
	According to~\cite[Lemma 11]{eldan2016}, we have
	\[
	\| f_R \phi \|_{L_2} = \| f_R \|_{L_2(\mu)} \geq \theta / \alpha .
	\]
\end{proposition}
The proofs of Propositions~\ref{prop:1},~\ref{prop:2}, and~\ref{prop:3} are shown in Appendix~\ref{app:useful}.

Based on the results above, we define the pair $(p,q)$
\[
p = \frac{\widehat{f\phi}}{\|f\phi\|_{L_2}} \quad \text{and} \quad q = \frac{\widehat{f_R\phi}}{\|f_R\phi\|_{L_2}},
\]
which satisfy the conditions of Propositions~\ref{prop:1} and~\ref{prop:2}. Thus, we have
\[
\begin{aligned}
	\| f - f_R \|_{L_2(\mu)} &= \| f\phi - f_R\phi \|_{L_2} = \| \widehat{f\phi} - \widehat{f_R\phi} \|_{L_2} = \| \left( \| f\phi \|_{L_2} \right) p - \left( \| f_R\phi \|_{L_2} \right) q \|_{L_2} \\
	& \geq \frac{1}{2} \|p - q\|_{L_2} \| f_R\phi \|_{L_2} \geq \frac{\theta}{2\alpha} \|p - q\|_{L_2} \geq \frac{\theta}{2\alpha} \sqrt{2(1-\langle p,q \rangle_{L_2})} \\
	& \geq \frac{\theta}{2\alpha} \sqrt{\max\{ \delta/2 - n_r e^{-2Cd},0\}} .
\end{aligned}
\]
Provided $n_r \leq C_1e^{2C_1d}$ and $C_1 = \min\{ \delta/4,C \}$, we have
\[
\| f - f_R \|_{L_2(\mu)} \geq \frac{\theta\sqrt{\delta}}{4\alpha} ,
\]
and
\[
\sqrt{\max\{ \delta/2 - n_r e^{-2Cd},0\}} \geq \sqrt{\delta/4}.
\]
This completes the proof.  $\hfill\square$

\section{Complete Proofs for Theorem~\ref{thm:minimas}}
We provided the detailed proofs for Theorem~\ref{thm:minimas}.

\subsection{Proof of Lemma~\ref{lemma:complex_affine}} \label{app:complex_affine}
For $i,j$ and $\rho \in \mathbb{R}$, let $\boldsymbol{v}_i = \rho\boldsymbol{\theta}_j+\boldsymbol{\theta}_i$ and $\boldsymbol{v}_j = (1-\rho)\boldsymbol{\theta}_j$, then we have
\begin{equation} \label{eq:rho_r}
	\begin{aligned}
		\begin{pmatrix}
			[\boldsymbol{v}_i]_R \\
			[\boldsymbol{v}_i]_I \\
			[\boldsymbol{v}_j]_R \\
			[\boldsymbol{v}_j]_I \\
		\end{pmatrix} &= \begin{pmatrix}
			[\boldsymbol{\theta}_i]_R \\
			[\boldsymbol{\theta}_i]_I \\
			0 \\
			0 \\
		\end{pmatrix} + \rho \begin{pmatrix}
			[\boldsymbol{\theta}_j]_R \\
			[\boldsymbol{\theta}_j]_I \\
			0 \\
			0 \\
		\end{pmatrix} + (1-\rho) \begin{pmatrix}
			0 \\
			0 \\
			[\boldsymbol{\theta}_j]_R \\
			[\boldsymbol{\theta}_j]_I \\
		\end{pmatrix} \\
		&= \left(\begin{array}{c c c c}
			1 & 0 & \rho & 0 \\
			0 & 1 & 0 & \rho \\
			0 & 0 & (1-\rho) & 0 \\
			0 & 0 & 0 & (1-\rho) \\
		\end{array} \right) \begin{pmatrix}
			[\boldsymbol{\theta}_i]_R \\
			[\boldsymbol{\theta}_i]_I \\
			[\boldsymbol{\theta}_j]_R \\
			[\boldsymbol{\theta}_j]_I \\
		\end{pmatrix} .
	\end{aligned}
\end{equation}
For the case $\rho = \rho_1 + \rho_2 \boldsymbol{i}\in\mathbb{C}$, one has
\begin{equation} \label{eq:rho_c}
	\begin{pmatrix}
		[\boldsymbol{v}_i]_R \\
		[\boldsymbol{v}_i]_I \\
		[\boldsymbol{v}_j]_R \\
		[\boldsymbol{v}_j]_I \\
	\end{pmatrix} = \left(\begin{array}{c c c c}
		1 & 0 & \rho_1 & -\rho_2 \\
		0 & 1 & \rho_2 & \rho_1 \\
		0 & 0 & (1-\rho_1) & \rho_2 \\
		0 & 0 & -\rho_2 & (1-\rho_1) \\
	\end{array} \right) \begin{pmatrix}
		[\boldsymbol{\theta}_i]_R \\
		[\boldsymbol{\theta}_i]_I \\
		[\boldsymbol{\theta}_j]_R \\
		[\boldsymbol{\theta}_j]_I \\
	\end{pmatrix} .
\end{equation}
So each linear mapping $\phi(\Theta;\rho,i,j)$ leads to a straight line in an 2-dimensional affine space. This completes the proof. $\hfill\square$

\subsection{Proof of Lemma~\ref{lemma:composition}} \label{app:composition}
The existence of real-valued equioutput transformations is proved by~\cite{chen1993}. The proof sketch of the complex-valued equioutput transformations are similar. There are three facts that (1) zReLU is even on the complex-valued domain, that is, $\text{zReLU}(-\boldsymbol{z}) = \text{zReLU}(\boldsymbol{z})$; (2) for $l\neq L$ and any $i,j,\rho_1,\rho_2$, there exists a pair of scalar parameters $\tilde{\alpha}_i$ and $\tilde{\alpha}_j$ in the next layer (i.e., $(l+1)$-th layer) such that
\[
L \left( \phi_2 \circ \phi_1(\Theta^l), \tilde{\alpha}_i,\tilde{\alpha}_j \right) = L(\Theta^l,\alpha_i,\alpha_j ) .
\]
(3) $( \tilde{\alpha}_i,\tilde{\alpha}_j )$ is led by a composition $\phi_1'\circ\phi'_2$ of other linear mappings for some $\rho'_1, \rho_2'$.

Next, we are going to prove facts (2) and (3). Provided the parameters $\{ \boldsymbol{\theta}_i, \boldsymbol{\theta}_j, \alpha_i, \alpha_j \}$ relative to neurons $i,j$, we have
\[
h(\boldsymbol{z}) = \alpha_i\sigma_{cr}( \boldsymbol{\theta}_i^T \boldsymbol{z} ) + \alpha_{j} \sigma_{cr}( \boldsymbol{\theta}_j^T \boldsymbol{z}) = \left(\begin{array}{c}
	1  \\
	\boldsymbol{i} \\
	1 \\
	\boldsymbol{i} \\
\end{array}\right)^{\top} \left(\begin{array}{cccc}
	1 & 0 & 0 & 0 \\
	0 & 1 & 0 & 0 \\
	0 & 0 & 1 & 0 \\
	0 & 0 & 0 & 1 \\
\end{array}\right) \begin{pmatrix}
	[\alpha_i^Tz_i]_R \\
	[\alpha_i^Tz_i]_I \\
	[\alpha_j^Tz_j]_R \\
	[\alpha_j^Tz_j]_I \\
\end{pmatrix} ,
\]
where $\sigma_{cr}$ denotes the element-wise activation, and
\[
\begin{pmatrix}
	[z_i]_R \\
	[z_i]_I \\
	[z_j]_R \\
	[z_j]_I \\
\end{pmatrix} = \sigma_{cr} \circ
\left(\begin{array}{cccc}
	1 & 0 & 0 & 0 \\
	0 & 1 & 0 & 0 \\
	0 & 0 & 1 & 0 \\
	0 & 0 & 0 & 1 \\
\end{array}\right) \begin{pmatrix}
	[\boldsymbol{\theta}_i^T\boldsymbol{x}]_R \\
	[\boldsymbol{\theta}_i^T\boldsymbol{x}]_I \\
	[\boldsymbol{\theta}_j^T\boldsymbol{x}]_R \\
	[\boldsymbol{\theta}_j^T\boldsymbol{x}]_I \\
\end{pmatrix} .
\]
Let $\rho_1 = \rho_{11}+\rho_{12}\boldsymbol{i}$, $\rho_2 = \rho_{21}+\rho_{22}\boldsymbol{i}$, $\rho_1' = \rho_{11}'+\rho_{12}'\boldsymbol{i}$, and $\rho_2' = \rho_{21}'+\rho_{22}'\boldsymbol{i}$. According to Eq.~\eqref{eq:rho_c}, one has
\begin{multline*}
	\tilde{h}(\boldsymbol{x}) = \tilde{\alpha}_{i}\sigma_{cr}( (\rho\boldsymbol{\theta}_j + \boldsymbol{\theta}_i)^T \boldsymbol{x} ) + \tilde{\alpha}_{j} \sigma_{cr}( (1-\rho)\boldsymbol{\theta}_j^T \boldsymbol{x}) \\
	= \left(\begin{array}{c}
		1  \\
		\boldsymbol{i} \\
		1 \\
		\boldsymbol{i} \\
	\end{array}\right)^{\top} \left(\begin{array}{cccc}
		1 & 0 & (1-\rho_{21}') & \rho_{22}' \\
		0 & 1 & -\rho_{22}' & (1-\rho_{21}') \\
		0 & 0 & \rho_{21}' & -\rho_{22}' \\
		0 & 0 & \rho_{22}' & \rho_{21}' \\
	\end{array}\right)  \left(\begin{array}{cccc}
		1 & 0 & \rho_{11}' & -\rho_{12}' \\
		0 & 1 & \rho_{12}' & \rho_{11}' \\
		0 & 0 & (1-\rho_{11}') & \rho_{12}' \\
		0 & 0 & -\rho_{12}' & (1-\rho_{11}') \\
	\end{array}\right) \begin{pmatrix}
		[\alpha_i^Tz_i']_R \\
		[\alpha_i^Tz_i']_I \\
		[\alpha_j^Tz_j']_R \\
		[\alpha_j^Tz_j']_I \\
	\end{pmatrix} ,
\end{multline*}
where
\[
\begin{pmatrix}
	[z_i']_R \\
	[z_i']_I \\
	[z_j']_R \\
	[z_j']_I \\
\end{pmatrix} = \sigma_{cr} \circ
\left(\begin{array}{cccc}
	1 & 0 & (1-\rho_{21}) & \rho_{22} \\
	0 & 1 & -\rho_{22} & (1-\rho_{21}) \\
	0 & 0 & \rho_{21} & -\rho_{22} \\
	0 & 0 & \rho_{22} & \rho_{21} \\
\end{array}\right)  \left(\begin{array}{cccc}
	1 & 0 & \rho_{11} & -\rho_{12} \\
	0 & 1 & \rho_{12} & \rho_{11} \\
	0 & 0 & (1-\rho_{11}) & \rho_{12} \\
	0 & 0 & -\rho_{12} & (1-\rho_{11}) \\
\end{array}\right) \begin{pmatrix}
	[\boldsymbol{\theta}_i^T\boldsymbol{x}]_R \\
	[\boldsymbol{\theta}_i^T\boldsymbol{x}]_I \\
	[\boldsymbol{\theta}_j^T\boldsymbol{x}]_R \\
	[\boldsymbol{\theta}_j^T\boldsymbol{x}]_I \\
\end{pmatrix} .
\]
Let $ \tilde{h}(\boldsymbol{x}) = h(\boldsymbol{x}) $, we obtain a semi-linear equation with eight free parameters. This equation has at least one solution, which completes the proof of facts (2) and (3). Lemma~\ref{lemma:composition} holds as desired. $\hfill\square$

\subsection{Proof of Lemma~\ref{lemma:manifold}} \label{app:manifold}
For real-valued neural networks, let $p$ and $q$ denote the neuron numbers of some adjacent layers. According to Section~\ref{sec:pre}, there are $pq$ parameters. Consider the following cases. (1) Both $p,q$ are even, that is, $p=2n$ and $q=2m$. We construct a complex-reaction network with two layers, where the first and second layers have $2n$ and $m$ neurons, respectively. Then the constructed complex-reaction network has $2(2n)m$ real-valued parameters. (2) When $p=2n+1$ and $q=2m$, we can construct a complex-reaction network with two layers, where the first and second layers have $p$ and $m$ neurons, respectively. Then the constructed complex-reaction network has $2pm$ real-valued parameters. (3) When $p=2n+1$ and $q=2m+1$, we can construct a complex-reaction network as follows: (3a) this network consist of two layers where the first and second layers have $n+1$ and $2m+1$ neurons; (3b) we force imaginary part of the last neuron of the first layer to be zero. Thus, half of the connection weights that link this neuron and $2m+1$ neurons of the second layer are useless. Then the constructed complex-reaction network has $2(n+1)(2m+1)-(2m+1)$ real-valued parameters.

Summing up the above, for any fully-connected feed-forward real-valued neural network, we can construct a complex-reaction network, which has the same parameter structure with the real-valued one. According to the differential manifold theory, the coordinate transformation of the complex manifold needs to satisfy the holomorphic condition. So the complex manifold led by the parameters of the constructed network is a subset of the real manifold led by that of the real-valued one, that is,
\[
\Omega_{CR}^l \subseteq \Omega_{R}^l,
\]
where the superscript $l$ denotes the $l$-th layer. According to Lemma~\ref{lemma:complex_group}, we can write the connection weights of each network as $\{ \Theta^1  \dots \Theta^L \}$, where $\Theta^l$ denotes the connection weight matrix concerning the $l$-th layer. Thus, the parameter space of each network is a direct product of the sub-manifolds led by each layer, that is,
\[
\Omega_{CR} = \Omega_{CR}^1 \times \dots \times \Omega_{CR}^L \quad \text{and} \quad \Omega_{R} = \Omega_{R}^1 \times \dots \times \Omega_{R}^L.
\]
Finally, we have
\[
\Omega_{CR} \subseteq \Omega_{R}.
\]
This completes the proof.   $\hfill\square$

\subsection{Proof of Proposition~\ref{prop:gradients}}
We begin our proof with some useful lemmas.
\begin{lemma} [Normalization Matrix]
	Let $\boldsymbol{w}, \boldsymbol{v} \in \mathbb{R}^{1\times n}$ with $\boldsymbol{w} = \gamma \boldsymbol{v}$ and $\| \boldsymbol{v} \|=1$. Define $\mathbf{S} = \mathbf{I}_{n\times n} - \boldsymbol{v}^T\boldsymbol{v}$, then we have
	\[
	(1)~ \mathbf{S} = \mathbf{I}_{n\times n} - \frac{\boldsymbol{w}^T \boldsymbol{w}}{\|\boldsymbol{w}\|_2^2}; \quad (2)~ \frac{\partial \boldsymbol{v}}{\partial \boldsymbol{w}} = \frac{\mathbf{S}}{\gamma}; \quad (3)~ \mathbf{S} \boldsymbol{w}^T = \mathbf{S} \boldsymbol{v}^T = \boldsymbol{0}; \quad (4)~ \mathbf{S}^2 = \mathbf{S}.
	\]
\end{lemma}
The detailed proof of Lemma~\ref{lemma:matrix} is given by Appendix~\ref{app:matrix}. 
\begin{lemma} [Weight Norms]
	During the gradient descent procedure, the change rate of $\| \gamma^l_j \|$ (i.e., weight norms) is the same for each layer.
\end{lemma}
Lemma~\ref{lemma:gamma} shows that $(\gamma^l_j)^2 = \| \mathbf{W}^l_j \|^2$ grows at a rate independent of the row $j$ and layer $l$. Thereby, using gradient descent to solve the optimization problem, there is no difference in the change rate of connection weights layer by layer. This result also holds for real-valued networks. The detailed proof is presented by Appendix~\ref{app:gamma}.

Here, we study the following minimization optimization problem of complex-reaction networks.
\[
\begin{aligned}
	& \min_{\mathbf{V}^l_j, \gamma_j^l} L( \mathbf{W};\mathbf{X} ) = \frac{1}{N} \sum_{n=1}^{N} \exp\left( -y_n [f( \mathbf{W}; \boldsymbol{x}_n )]_R \right) , \\
	& s.t.~~ \mathbf{W}^l_j = \gamma^l_j~ \mathbf{V}^l_j, ~~\|\mathbf{V}^l_i\| =1.
\end{aligned}
\]
Solving this problem by the standard gradient descent, the optimization procedure concerning $\mathbf{W}^l$ induces the following dynamical system:
\[
\frac{d \mathbf{W}^l}{d t} = - \frac{\partial L}{\partial \mathbf{W}^l} =\frac{1}{N} \sum_{n=1}^{N} \exp\left( -y_n [f( \mathbf{W}; \boldsymbol{x}_n )]_R \right) y_n \frac{\partial [f( \mathbf{W}; \boldsymbol{x}_n )]_R}{\partial \mathbf{W}^l} .
\]
Based on the Euler's theorem for homogeneous functions~\cite{bronshtein2013}, we have
\[
[f( \mathbf{W}; \mathbf{X} )]_R = \mathbf{W}^l_i~ \frac{\partial [f( \mathbf{W}; \mathbf{X} )]_R}{\partial \mathbf{W}^l_i} = \mathbf{W}^l_j \cdot \left[ \frac{\partial [f( \mathbf{W};\mathbf{X} )]_R }{\partial \mathbf{W}^l_j} + \frac{\partial [f( \mathbf{W};\mathbf{X} )]_R }{\partial \mathbf{W}^l_j} \boldsymbol{i} \right]
\]
and
\[
\frac{\partial f(\mathbf{W}^l)}{\partial \mathbf{W}^l} \propto \frac{\partial f(\mathbf{V}^l)}{\partial \mathbf{V}^l}.
\]
The formula above implies that the standard gradient descents of non-normalized and normalized complex-reaction networks are proportional. Thus, the gradient descent procedure with weight normalization induces the following dynamical systems
\begin{equation}  \label{eq:uncon_v}
	\left\{\begin{aligned}
		\frac{d \gamma^l}{d t} &= - \frac{\partial L}{\partial \gamma^l} = \sum_{n=1}^{N} \exp\left( -y_n [f( \mathbf{W}; \boldsymbol{x}_n )]_R \right) y_n \frac{\partial [f( \mathbf{W}; \boldsymbol{x}_n )]_R}{\partial \gamma^l} ,\\
		\frac{d \mathbf{V}^l}{d t} &= - \frac{\partial L}{\partial \mathbf{V}^l} = \sum_{n=1}^{N} \exp\left( -y_n [f( \mathbf{W}; \boldsymbol{x}_n )]_R \right) y_n \frac{\partial [f( \mathbf{W}; \boldsymbol{x}_n )]_R}{\partial \mathbf{V}^l} .\\
	\end{aligned}\right.
\end{equation}
Define $\boldsymbol{w}_j = ( [\mathbf{W}^l_j]_R, -[\mathbf{W}^l_j]_I )$\footnote{If $[\mathbf{W}^l_j]_R$ has $d$ elements, $\boldsymbol{w}_j$ is a $2d$-dimensional row vector.}, $\boldsymbol{v}_j = ( [\mathbf{V}^l_j]_R, -[\mathbf{V}^l_j]_I )$, and $\mathbf{S}_j = \mathbf{I} - \boldsymbol{v}_j^T\boldsymbol{v}_j$. According to Lemma~\ref{lemma:matrix}, one has
\[
\mathbf{S}_j = \mathbf{I} - \frac{\boldsymbol{w}_j^T \boldsymbol{w}_j}{\|\boldsymbol{w}_j\|_2^2} ~~\text{and}~~ \mathbf{S}_j \boldsymbol{w}_j^T = \mathbf{S}_j \boldsymbol{v}_j^T = \boldsymbol{0}.
\]
Thus, we have
\[
\frac{d \gamma^l_j}{d t} = \frac{d \|\boldsymbol{w}_j \|}{d t} = \boldsymbol{v}_j \left( \frac{d \boldsymbol{w}_j}{d t} \right)^T ~~\text{and}~~ \frac{d \boldsymbol{v}_j}{d t} = \frac{\mathbf{S}_j}{\gamma^l_j} \left( \frac{d \boldsymbol{w}_j}{d t} \right)^T.
\]
So Eq.~\eqref{eq:uncon_v} becomes
\[
\left\{ \begin{aligned}
	\frac{d \gamma^l_j}{d t} &= \frac{\eta}{\gamma^l_j} \frac{1}{N} \sum_{n=1}^N y_n \exp\left( -y_n [f( \mathbf{W}; \boldsymbol{x}_n )]_R \right) \boldsymbol{v}_j \boldsymbol{u}_j^T ,\\
	\frac{d \boldsymbol{v}_j}{d t} &= \frac{\eta}{\left(\gamma^l\right)^2} \frac{1}{N} \sum_{n=1}^N y_n \exp\left( -y_n [f( \mathbf{W}; \boldsymbol{x}_n )]_R \right) \left( \boldsymbol{u}_j - \boldsymbol{v}_j [f( \mathbf{V}; \boldsymbol{x}_n )]_R \right) ,\\
\end{aligned} \right.
\]
where
\[
\boldsymbol{u}_j = \left( \frac{\partial [f( \mathbf{V}; \boldsymbol{x}_n )]_R}{\partial \left[ \mathbf{V}_j^l \right]_R }, - \frac{\partial [f( \mathbf{V}; \boldsymbol{x}_n )]_R}{\partial \left[ \mathbf{V}_j^l \right]_I } \right) .
\]
Due to
\[
\boldsymbol{v}_j \boldsymbol{u}_j^T = [\mathbf{V}_j^l]_R \left[ \frac{\partial [f( \mathbf{V}; \boldsymbol{x}_n )]_R}{\partial \left[ \mathbf{V}_j^l \right]_R } \right]^T + [\mathbf{V}_j^l]_I \left[ \frac{\partial [f( \mathbf{V}; \boldsymbol{x}_n )]_R}{\partial \left[ \mathbf{V}_j^l \right]_I } \right] ^T = [f( \mathbf{V}; \boldsymbol{x}_n )]_R ,
\]
we have
\begin{equation} \label{eq:S}
	\left\{ \begin{aligned}
		\frac{d \gamma^l_j}{d t} &= \frac{\eta}{\gamma^l_j} \frac{1}{N} \sum_{n=1}^N y_n \exp\left( -y_n [f( \mathbf{W}; \boldsymbol{x}_n )]_R \right) [f( \mathbf{V}; \boldsymbol{x}_n )]_R ,\\
		\frac{d \boldsymbol{v}_j}{d t} &= \frac{\eta}{\left(\gamma^l_j\right)^2} \frac{1}{N} \sum_{n=1}^N y_n \exp\left( -y_n [f( \mathbf{W}; \boldsymbol{x}_n )]_R \right) \left( \boldsymbol{u}_j - \boldsymbol{v}_j [f( \mathbf{V}; \boldsymbol{x}_n )]_R \right) .\\
	\end{aligned} \right.
\end{equation}

Let $\mathbf{I}_{\mathbb{C}} = ( \mathbf{I}_{d \times d}  ,~-\boldsymbol{i}~ \mathbf{I}_{d \times d} )^T $ and multiply it by Eq.~\eqref{eq:S}. Thus, we can obtain the gradient descent dynamics concerning the normalized network
\[
\left\{ \begin{aligned}
	\frac{d \gamma^l_j}{d t} &= \frac{\eta}{\gamma^l_j} \frac{1}{N} \sum_{n=1}^N y_n \exp\left( -y_n [f( \mathbf{W}; \boldsymbol{x}_n )]_R \right) [f( \mathbf{V}; \boldsymbol{x}_n )]_R ,\\	\frac{d \mathbf{V}^l_j}{d t} &=
	\frac{d \boldsymbol{v}_j}{d t} \mathbf{I}_{\mathbb{C}} = \frac{\eta}{\left(\gamma^l_j\right)^2} \frac{1}{N} \sum_{n=1}^N y_n \exp\left( -y_n [f( \mathbf{W}; \boldsymbol{x}_n )]_R \right) \Delta_j ,\\
\end{aligned} \right.
\]
where
\[
\Delta_j = \left( \frac{\partial [f( \mathbf{V}; \boldsymbol{x}_n )]_R}{\partial \left[ \mathbf{V}_j^l \right]_R } - [\mathbf{V}_j^l]_R [f( \mathbf{V}; \boldsymbol{x}_n )]_R \right) + \left(  \frac{\partial [f( \mathbf{V}; \boldsymbol{x}_n )]_I}{\partial \left[ \mathbf{V}_j^l \right]_I } - [\mathbf{V}_j^l]_I [f( \mathbf{V}; \boldsymbol{x}_n )]_R \right) \boldsymbol{i} .
\]
This completes the proof. $\hfill\square$

\subsection{Proof of Proposition~\ref{lemma:gradients_r}}
Consider a real-valued neural network $f_R:\mathbb{R}^{2d} \to \{-1,+1\}$ with ReLU activation and weight normalization $\mathbf{P}^l_j = \gamma_j^l \mathbf{Q}^l_j$, where $\gamma_j^l \in \mathbb{R}^+$ and $\| \mathbf{Q}^l_j \|=1$. Given a training set $\{\boldsymbol{x}_n,y_n\}_{n=1}^{N}$ with $\mathbf{X} = \{\boldsymbol{x}_n\}_{n=1}^{N}$, we employ standard gradient descents to minimize the empirical exponential loss.

First, we should introduce some necessary facts. The ReLU function has the real-valued homomorphism property~\cite{poggio2020}, that is, for any $x \in \mathbb{R}$ and $\alpha \geq 0$, the equation holds
\[
\sigma_r(\alpha x) = \alpha \sigma_r(x) .
\]
Thus, we have
\[
\sigma_r(x) = \frac{\partial \sigma_r(x)}{\partial x} x ,
\]
and then,
\[
f_R(\mathbf{P};\mathbf{X}) = \mathbf{P}_j^l \left( \frac{\partial f_R(\mathbf{P};\mathbf{X})}{\partial \mathbf{P}_j^l} \right)^T ~~\text{and}~~ f_R(\mathbf{Q};\mathbf{X}) = \mathbf{Q}_j^l \left( \frac{\partial f_R(\mathbf{Q};\mathbf{X})}{\partial \mathbf{Q}_j^l} \right)^T ,~~\text{respectively}.
\]
The optimization procedure concerning $\mathbf{P}^l$ is led by the following dynamical systems
\[
\frac{d\mathbf{P}^l}{dt} = -\frac{\partial L}{\partial \mathbf{P}^l} = \frac{1}{N} \sum_{n=1}^{N} \exp\left( -y_nf_R(\mathbf{P};\boldsymbol{x}_n) \right) y_n \frac{\partial f_R(\mathbf{P};\boldsymbol{x}_n)}{\partial \mathbf{P}^l} .
\]
The gradient descent procedure with weight normalization induces the following dynamical systems
\begin{equation}  \label{eq:prop2}
	\left\{\begin{aligned}
		\frac{d \gamma^l}{d t} &= - \frac{\partial L}{\partial \gamma^l} = \sum_{n=1}^{N} \exp\left( -y_nf_R(\mathbf{P};\boldsymbol{x}_n) \right) y_n \frac{\partial f_R( \mathbf{P}; \boldsymbol{x}_n )}{\partial \gamma^l} ,\\
		\frac{d \mathbf{V}^l}{d t} &= - \frac{\partial L}{\partial \mathbf{Q}^l} = \sum_{n=1}^{N} \exp\left( -y_nf_R(\mathbf{P};\boldsymbol{x}_n) \right) y_n \frac{\partial f_R( \mathbf{P}; \boldsymbol{x}_n )}{\partial \mathbf{Q}^l} .\\
	\end{aligned}\right.
\end{equation}
Similar to the proof of Proposition~\ref{prop:gradients}, we use the vectorized representation, i.e., denote $\boldsymbol{w}_j = \mathbf{P}^l_j$ and $\boldsymbol{v}_j =\mathbf{Q}^l_j$. Define a matrix
\[
\mathbf{S}_j = \mathbf{I} - \boldsymbol{v}_j^T\boldsymbol{v}_j = \mathbf{I} - \frac{\boldsymbol{w}_j^T\boldsymbol{w}_j}{\| \boldsymbol{w}_j^T\boldsymbol{w}_j\|} .
\]
According to Lemma~\ref{lemma:matrix}, we have
\[
\frac{d \gamma^l_j}{d t} = \frac{d \|\boldsymbol{w}_j \|}{d t} = \boldsymbol{v}_j \left( \frac{d \boldsymbol{w}_j}{d t} \right)^T ~~\text{and}~~ \frac{d \boldsymbol{v}_j}{d t} = \frac{\mathbf{S}_j}{\gamma^l_j} \left( \frac{d \boldsymbol{w}_j}{d t} \right)^T, ~~\text{respectively} .
\]
Thus, Eq.~\eqref{eq:prop2} becomes
\[
\left\{ \begin{aligned}
	\frac{d \gamma^l_j}{d t} &= \frac{\eta}{\gamma^l_j} \frac{1}{N} \sum_{n=1}^N y_n \exp\left( -y_nf_R(\mathbf{P};\boldsymbol{x}_n) \right) \mathbf{Q}_j^l \left( \frac{\partial f_R(\mathbf{Q};\mathbf{X})}{\partial \mathbf{Q}_j^l} \right)^T \\
	& = \frac{\eta}{\gamma^l_j} \frac{1}{N} \sum_{n=1}^N y_n \exp\left( -y_nf_R(\mathbf{P};\boldsymbol{x}_n) \right) \mathbf{Q}_j^l \left( \frac{\partial f_R(\mathbf{Q};\mathbf{X})}{\partial \mathbf{Q}_j^l} \right)^T ,\\
	\frac{d \boldsymbol{v}_j}{d t} &= \frac{\eta}{\left(\gamma^l\right)^2} \frac{1}{N} \sum_{n=1}^N y_n \exp\left( -y_nf_R(\mathbf{P};\boldsymbol{x}_n) \right) \mathbf{S}_j \frac{\partial f_R(\mathbf{Q};\mathbf{X})}{\partial \mathbf{Q}_j^l} \\
	&= \frac{\eta}{\left(\gamma^l\right)^2} \frac{1}{N} \sum_{n=1}^N y_n \exp\left( -y_nf_R(\mathbf{P};\boldsymbol{x}_n) \right) \left( \frac{\partial f_R(\mathbf{Q};\mathbf{X})}{\partial \mathbf{Q}_j^l} - \mathbf{Q}_j^l f_R( \mathbf{Q}; \boldsymbol{x}_n ) \right) ,\\
\end{aligned} \right.
\]
where $\eta$ is a strictly positive constant relative to $\gamma_j^l$, which satisfies
\[
f(\mathbf{P}; \mathbf{X}) = \eta f(\mathbf{Q}; \mathbf{X}) ~~\text{and}~~ \frac{\partial f(\mathbf{P}^l_j)}{\partial \mathbf{P}^l_j}  = \frac{\eta}{\gamma_j^l} \frac{\partial f(\mathbf{Q}^l_j)}{\partial \mathbf{Q}^l_j}, ~~\text{respectively} .
\]
This completes the proof. $\hfill\square$

\section{Complete Proofs for Useful Lemmas} \label{app:useful}
This section complete the proofs of some useful lemmas in Appendix.
\subsection{Proof of Lemma~\ref{lemma:matrix}} \label{app:matrix}
Let $\boldsymbol{w}=(w_1,\dots,w_n), \boldsymbol{v}=(v_1,\dots,v_n)$. Since $\boldsymbol{w} = \gamma \boldsymbol{v}$ and $\| \boldsymbol{v} \|=1$, we have
\[
\| \boldsymbol{v} \| =  \sqrt{(v_1)^2 + \dots + (v_n)^2} = 1 \quad\text{and}\quad w_i = \gamma v_i \quad \text{for any} \quad i \in [n] .
\]
Let
\[
\mathbf{S} = \mathbf{I}_{n\times n} - \boldsymbol{v}^T\boldsymbol{v} = \begin{pmatrix}
	1-(v_1)^2 & -v_1v_2 & \cdots & -v_1v_n \\
	-v_2v_1 & 1-(v_2)^2 & \cdots & -v_2v_n \\
	\vdots & \vdots & \ddots & \vdots \\
	-v_nv_1 & -v_nv_2 & \cdots & 1-(v_n)^2
\end{pmatrix} .
\]
On the other hand, we have
\[
\mathbf{I}_{n\times n} - \frac{\boldsymbol{w}^T \boldsymbol{w}}{\|\boldsymbol{w}\|_2^2} = \mathbf{I}_{n\times n} - \frac{1}{\|\boldsymbol{w}\|_2^2} \begin{pmatrix}
	(w_1)^2 & w_1w_2 & \cdots & w_1w_n \\
	w_2w_1 & (w_2)^2 & \cdots & w_2w_n \\
	\vdots & \vdots & \ddots & \vdots \\
	w_nw_1 & w_nw_2 & \cdots & (w_n)^2
\end{pmatrix} .
\]
For $i,j \in[n]$, one has
\[
1 - \frac{(w_i)^2}{(w_1)^2 + \dots + (w_n)^2} = \frac{\sum_{k\neq i}(w_k)^2}{\sum_{k}(w_k)^2} = \frac{\sum_{k\neq i}(v_k)^2}{\sum_{k}(v_k)^2} = 1 - \frac{(v_i)^2}{(v_1)^2 + \dots + (v_n)^2} = 1 - (v_i)^2 ,
\]
and
\[
- \frac{w_iw_j}{(w_1)^2 + \dots + (w_n)^2} = - \frac{v_iv_j}{(v_1)^2 + \dots + (v_n)^2} = - v_iv_j .
\]
Thus, we have
\[
\mathbf{S} = \mathbf{I}_{n\times n} - \frac{\boldsymbol{w}^T \boldsymbol{w}}{\|\boldsymbol{w}\|_2^2} .
\]
Consider the partial derivative of $\boldsymbol{v}$ with respect to  $\boldsymbol{w}$
\[
\frac{\partial \boldsymbol{v}}{\partial \boldsymbol{w}} =  \begin{pmatrix}
	\partial v_1 / \partial w_1 & \partial v_1 / \partial w_2 & \cdots & \partial v_1 / \partial w_n \\
	\partial v_2 / \partial w_1 & \partial v_2 / \partial w_2 & \cdots & \partial v_2 / \partial w_n \\
	\vdots & \vdots & \ddots & \vdots \\
	\partial v_n / \partial w_1 & \partial v_n / \partial w_2 & \cdots & \partial v_n / \partial w_n
\end{pmatrix} .
\]
Thus, we have
\[
\frac{\partial \boldsymbol{v}}{\partial \boldsymbol{w}} = \frac{\mathbf{S}}{\gamma} .
\]
It is easily to verify that $0$ is an eigenvalue of the matrix $\mathbf{S}$ and $\boldsymbol{v}$ is the corresponding eigenvector. So we have
\[
\mathbf{S} \boldsymbol{w}^T = \mathbf{S} \boldsymbol{v}^T = \boldsymbol{0} .
\]
Let $\mathbf{S}_i$ denote the $i$-th row vector of matrix $\mathbf{S}$. Thus, we have
\[
\mathbf{S}_i \left( \mathbf{S}_i \right)^{\top} = \sum_{k\neq i} (v_iv_k)^2 + \left( 1-(v_i)^2 \right)^2 = \sum_{k\neq i} (v_iv_k)^2 + \left( \sum_{k\neq i} (v_k)^2 \right)^2 = 1 - (v_i)^2 ,
\]
and
\[
\begin{aligned}
	\mathbf{S}_i \left( \mathbf{S}_j \right)^{\top} &= \sum_{k\neq i,j} (v_iv_j)(v_k)^2 + \left( 1-(v_i)^2 \right)v_jv_i + v_iv_j\left( 1-(v_j)^2 \right) \\
	& = \sum_{k\neq i,j} (v_iv_j)(v_k)^2 + v_iv_j \left( 2\sum_{k\neq i,j} (v_k)^2 + (v_i)^2 + (v_j)^2 \right) = - v_iv_j.
\end{aligned} 
\]
Thus, we have
\[
\mathbf{S}^2 = \mathbf{S}.
\]
This completes the proof.  $\hfill\square$

\subsection{Proof of Lemma~\ref{lemma:gamma}} \label{app:gamma}
Observing the change rate of $\gamma^l_j$ in Eq.~\eqref{eq:complex_gradients}, we have
\[
\frac{d \left( \gamma^l_j \right)^2}{d t} = 2~ \gamma^l_j \frac{d \gamma^l_j}{d t}  = 2 \frac{\eta}{N}\sum_{n=1}^{N} \exp\left( -y_n [f( \mathbf{W}; \boldsymbol{x}_n )]_R \right)~ [f( \mathbf{V}; \boldsymbol{x}_n )]_R.
\]
Thus, we have
\[
\left\| \frac{d \mathbf{W}^l_j}{d t} \right\| = \frac{\partial \|\mathbf{W}^l_j\|}{\partial \mathbf{W}^l_j} \frac{d \mathbf{W}^l_j}{d t} ,
\]
and then
\[
\left\| \frac{d \mathbf{W}^l_j}{d t} \right\|^2 = \frac{2}{N}\sum_{n=1}^{N} \exp\left( -y_n [f( \mathbf{W}; \boldsymbol{x}_n )]_R \right) [f( \mathbf{W}; \boldsymbol{x}_n )]_R.
\]
So the change rate of $\|\mathbf{W}^l_j\|^2$ is independent of the layer index $l$ and row index $j$. Finally, it is easily to verify that these results above also hold for Eq.~\eqref{eq:real_gradients}. $\hfill\square$

\end{document}